%% file: main.tex
\documentclass[11pt]{article}

\input{header}

\usepackage{etoc}
\usepackage{pdflscape}
\usepackage{etoolbox}
\usepackage{adjustbox}
\usepackage{algorithm}
\usepackage{algpseudocode}
\usepackage{float}

\newcommand{\tit}{Pre-Trained Language Models as Zero-Shot Tools for Social Science Research\thanks{I thank Ryan Enos for providing access to the TargetSmart voter files used in this paper.}}
\title{{\bf\tit}}

\author{Noah Dasanaike\thanks{PhD Candidate, Department of Government, Harvard University.}}
\date{\today}

\begin{document}

\maketitle
\etocdepthtag.toc{mainpaper}

\begin{abstract}
Record linkage and demographic inference from names are routine demands in social science. Most approaches to these problems require labeled training data that are difficult or costly to acquire. I show that pre-trained language models perform both tasks accurately without requiring task-specific training. For record linkage, I introduce EnsembleLink, an algorithm that uses pre-trained embeddings alongside string-based scorers to retrieve a short list of candidates and then identify the likeliest matches. EnsembleLink matches or exceeds the performance of most prior approaches while generalizing across scripts and formats that those methods are unable to handle. For ethnicity inference from names, I show that large language models (LLMs) achieve up to 83.8\% accuracy on U.S.\ voter files, compared to 78.0\% for name-supplemented BISG. Open-weight models offer reproducible implementations that approach these models in accuracy. Moreover, LLMs are able to infer group membership for religious, caste-based, and other ethnic group categories outside of the United States, and reduce bias in downstream estimates of geographic racial gaps. Where pre-training encodes the world knowledge required by each task, task-specific labels add less information.

\bigskip
\noindent {\bf Keywords:} record linkage, ethnicity inference, large language models, embeddings, Bayesian Improved Surname Geocoding
\end{abstract}
\vfill
{\small
\noindent\textbf{AI Disclosure:} During revision, the author used Claude Opus 5.5 (Anthropic) for proofreading, copyediting, LaTeX formatting, validating analysis code, and preparing replication documentation and some appendix materials. All changes were manually reviewed.\\
\par}

\newpage

\section{Introduction}

Empirical social scientists regularly work with messy string-based data in order to facilitate downstream research, including joining different datasets on shared identifiers (record linkage) and inferring demographic characteristics from names (ethnicity\footnote{I use ``ethnicity'' hereafter to refer to race, caste, religious sect, and related ascriptive categories.} inference, or ``surname geocoding''). While prior methods have been developed for accomplishing these tasks, they are often limited in the information that they can use. In practice, many record linkage approaches rely on literal string comparison between potential pairs, which handles misspellings well but is less capable of recognizing matches written in different scripts, entered in the wrong fields, or abbreviated or translated \citep{enamorado2019using}. 

The most common method for ethnicity inference, and specifically racial inference, in the United States is Bayesian Improved Surname Geocoding (BISG). BISG requires race-surname frequency tables from the U.S.\ Census \citep{imai2016improving} and can predict group membership only for the coarse racial categories defined by these tables. It also requires data on the geographic distribution of different racial groups. Given these strict data requirements, BISG cannot be applied in most countries or to categories beyond race, such as detailed ethnicity, religion, or caste, where no such surname frequency tables or geographic distributions exist. Even within the United States, BISG produces systematic error that biases downstream analysis, for instance disproportionately labeling Black voters in wealthy neighborhoods as White \citep{argyle2024misclassification}.

Although the two tasks may appear different, each reflects inference of an unobserved property of a textual record: which record(s) in one dataset a record in another refers to, and to which racial or ethnic group a given name likely belongs. Methods that extend beyond literal string overlap between potential matches, both of records to one another and of individuals to ethnic classifications, require labeled training data or population-specific name tables that are difficult or costly to acquire.

I show that pre-trained language models let researchers relax both constraints, and can be used for record linkage and ethnicity inference without requiring example data or additional fine-tuning. For record linkage, I introduce EnsembleLink, an algorithm that uses pre-trained text embeddings alongside string-based scorers to retrieve short candidate lists of matches for a given record, and then identifies the likeliest match from that set. EnsembleLink requires no labeled training data, performs as well as fastLink \citep{enamorado2019using} and Splink \citep{linacre2022splink} on clean administrative records, and performs better on records that are missed by these methods.

Beyond clean administrative records, EnsembleLink links multi-field listings, census occupations, and paragraphs. When linking names of people, organizations, and places written in Arabic, Chinese, Hindi, Japanese, Russian, or Spanish to their English translations, EnsembleLink also outperforms fuzzylink \citep{ornstein2025fuzzylink}, which labels pairs with proprietary large language models. When fuzzylink is trained on validation data, as the other methods are, rather than labeling the test data directly, EnsembleLink outperforms it on most tasks and performs comparably on the rest. EnsembleLink moreover scales to a million records without a blocking key (a shared field, such as state or surname initial, that restricts which records are compared). Blocking requirements have limited the scope of previous linkage methods \citep{enamorado2019using, linacre2022splink}.

For the task of inferring race or ethnicity from surnames, I show that proprietary and open-weight large language models (LLMs), prompted with a full name and geography, achieve up to 83.8\% accuracy on U.S.\ voter files, compared to 78.0\% for name-supplemented BISG. Without geography, where BISG cannot be estimated at all, their accuracy is nearly unchanged. I demonstrate that the same approach extends, without additional labels, to religious sects in Lebanon, caste in India, and country-specific ethnic groups (such as tribes in Uganda) in six countries with full-count voter rolls. 

Across a panel of thirty-five models, several open-weight models outperform standard BISG, the two DeepSeek-V4 models come within one point of name-supplemented BISG in every state when reasoning is enabled, and all offer replicability with fixed local weights \citep{spirling2023open}. I also obtain probabilities that are well calibrated in testing by training a small local classifier on the LLM-generated labels, and I show that LLM-based inference reduces the neighborhood-income bias that BISG introduces into downstream estimates of racial gaps. Finally, I propose an agentic workflow for validating the outputs of both tasks.

\section{Related Work}

For record linkage, the canonical approach is the probabilistic framework of \citet{fellegi1969theory}, which involves comparing each pair of records field by field and treating agreement on a field as evidence of a match. For names and other textual information, the approach involves scoring how lexically similar (e.g., Jaro--Winkler) the two strings are. ``Robert'' and ``Roberta'', for instance, despite referring in all likelihood to two different people, have a near-perfect string similarity (0.97), while ``Robert'' and ``Bob'' have a low one (0.50). The model estimates, without labels, which fields are more informative of a match than others. For instance, two records of the same person almost always share a birth year while two different people rarely do, and the model therefore learns that agreement on birth year is strong evidence of a match. 

The model then combines this information with similarity across fields to provide a match probability for each potential pair. Because the comparison relies on string similarity, names that are written in different scripts, abbreviated, or entered in the wrong field register as disagreements, even when both records refer to the same entity. For the same reasons, the approach typically performs much worse on records corresponding to entities other than people. For people, names are usually similar enough across sources, and metadata such as date of birth and sex are informative. The most widely used implementations of \citet{fellegi1969theory}, fastLink \citep{enamorado2019using} in R and Splink \citep{linacre2022splink} in Python, fit the Fellegi--Sunter weights to the data using expectation-maximization.

A second and more recent set of approaches treats record linkage as a binary classification task and uses active learning to extend a small set of training labels to larger datasets. The fuzzylink package \citep{ornstein2025fuzzylink} uses proprietary large language models to generate training labels, while \citet{kaufman2022ppp} learn the match function from labeled pairs by training an ensemble over a hand-coded set of classical string-distance features. Both of these methods depend on costly task-specific annotation, whether they compare records by string distance or by pre-trained embeddings.

Prior work on name-based ethnicity inference largely follows two approaches: Bayesian methods that combine race-surname distributions with geographic priors \citep{imai2016improving, dasanaike2026embedding, chasalow2026probabilistic}, and machine learning classifiers that train on labeled corpora. BISG \citep{imai2016improving} is the prevailing method used in the United States, where the U.S. Census Bureau publishes detailed race-surname frequency tables and various data sources offer granular geographic distributions of racial groups. The fBISG method \citep{imai2022addressing} improves upon BISG by adding a correction for measurement error and generalizing the method to include first names, and \citet{greengard2025calibrated} add a raking-based calibration. 

Alternatives to BISG largely comprise supervised machine learning classifiers such as Ethnicolr \citep{chintalapati2018predicting}, Rethnicity \citep{xie2022rethnicity}, and NamePrism \citep{ye2017nationality}, which achieve accuracy roughly comparable to BISG within the United States. Like BISG, however, they require labeled data for each population and ethnic category, which are in many settings unavailable.

\section{Why Do Pre-Trained Models Work?}
\label{sec:why}

Record linkage and name-based inference of ethnicity are both \emph{world knowledge} tasks. They require knowing, respectively, how entities relate to one another, and what names mean and which groups carry them. Existing methods learn this knowledge anew for each task, either from labeled training examples or from population-specific name tables. Pre-trained language models, conversely, acquire this information during pre-training on trillions of tokens of web text, books, news, and other documents of diverse origin. They should therefore already hold knowledge that can be applied to both tasks without additional labels.

If this prediction holds, several results should follow. For record linkage, matching based on these pre-trained language models should succeed even where a query and its true match share few or no characters, because training corpora commonly refer to the same entity under very different strings. For ethnicity inference, first and middle names, as well as names that are uncommon and thus not represented in frequency tables (the U.S. Census Bureau publishes only names occurring at least 100 times), should prove informative of group membership, because training corpora place full names alongside ethnic, religious, and national identifiers. 

Moreover, ethnicity inference should extend to non-Western contexts without additional training, because the corpora on which language models are trained are generally diverse in scope. However, models should classify names more accurately the more distinctive a name is to a group, and less accurately where naming conventions have converged across groups in a given context. This is not necessarily a limitation of the models but a limit on the information that a name (or name and geography) carries. For both ethnicity inference and record linkage, task-specific labels should add less to sufficiently capable models where the requisite knowledge is already found in their weights. Fine-tuning should help most where the models lack this knowledge, such as in small-parameter language models or tasks that depend on context that is particular to a given dataset. 

\section{Record Linkage}

\subsection{EnsembleLink}

While Fellegi--Sunter methods treat record linkage as a classification problem, EnsembleLink is a retrieve-and-rerank algorithm that treats it as a retrieval problem. For each record in one dataset (the query), EnsembleLink ranks the records in another dataset (the target) by how likely they are to refer to the same entity. EnsembleLink runs without labeled training data in two stages. For each query, the algorithm first narrows the target data to a short list of candidates, and then identifies the likeliest match(es) from that short list using an ensemble of scorers. 

The algorithm proceeds as follows (Algorithm~\ref{alg:main}). Each record is first concatenated into a single string. For records with more than one field, each value is preceded by the name of its field (e.g., \texttt{first\_name=JOHN | last\_name=DOE}). These field names let the model distinguish types of information. Records with a single field are used as-is, while for multi-field queries, the embedding model is provided a short instruction to retrieve the record that refers to the same entity.

In the first stage, two retrievers run in parallel. A dense retriever encodes every record with a pre-trained embedding model, which places strings with similar meanings close together in high-dimensional space. A sparse retriever compares short character sequences (n-grams) of the query and each target record. In the implementation here, names in non-Latin scripts are first transliterated, although the EnsembleLink results are robust to removing this step. Each retriever returns its top $k$ candidates (here I use $k=30$), and the union of these two lists (Appendix~\ref{app:retrieval}) is passed on to the second stage.

The second stage identifies the likeliest match or matches from this candidate list. Because no single approach is reliable a priori, the algorithm scores each candidate using an ensemble of experts. For the present implementation of EnsembleLink, I use four experts: a reranker that averages two cross-encoders, each of which passes each query and candidate together into a Transformer network; the embedding similarity, normalized so that records that happen to be close to many queries are not over-rewarded \citep{conneau2018word}; and two measures of character-level similarity.

Since there are no labels from which to learn how reliable each expert is in a given setting, and therefore how much each expert should be weighted, EnsembleLink infers reliability from agreement. Specifically, each expert is weighted by the squared share of queries on which its top candidate coincides with the consensus of all experts. An expert that systematically disagrees with the other experts, such as a cross-encoder on a purely lexical task, therefore receives little weight. For this reason, removing the transliteration step costs little accuracy, even when joining records in different scripts. Without transliteration, the string similarity scorers return near-random values on a task with many cross-script pairs, and these scores are in turn down-weighted (Appendix Table~\ref{tab:translit}).

By default, the final match for each query is the candidate with the highest weighted sum of standardized scores (I compare this rule with alternatives in Appendix~\ref{app:linkage_ablation}). These scores can also be thresholded to leave queries without a true match unmatched, and a one-to-one assignment can be imposed when a task requires it (Appendix~\ref{app:onetoone}). They also serve as a measure of confidence which, after calibration on a small labeled sample, can be interpreted probabilistically (Appendix~\ref{app:calibration}).

\begin{algorithm}[H]
\caption{EnsembleLink: ensemble retrieval and unsupervised agreement fusion}
\label{alg:main}
\begin{algorithmic}[1]
\Require Queries $Q$, target records $\mathcal{C}$, experts $e_1,\dots,e_E$, retrieval depth $k$
\Ensure A matched record for each query
\For{each query $q \in Q$}
    \State $\mathcal{R}_q \gets \text{TopK}_{\text{dense}}(q, \mathcal{C}, k) \cup \text{TopK}_{\text{sparse}}(q, \mathcal{C}, k)$ \Comment{candidate pool}
    \For{each expert $e$}
        \State $z_e(q,\cdot) \gets$ standardize $e$'s scores over $\mathcal{R}_q$
        \State $\hat{c}_e(q) \gets \arg\max_{c \in \mathcal{R}_q} z_e(q,c)$, or no vote if $z_e(q,\cdot)$ is constant \Comment{expert $e$'s top pick}
    \EndFor
    \State $\bar{c}(q) \gets \operatorname{mode}_e \hat{c}_e(q)$ over the experts that vote \Comment{cross-expert consensus}
\EndFor
\For{each expert $e$}
    \State $w_e \gets \big( |Q|^{-1} \sum_{q \in Q} \mathbf{1}[\hat{c}_e(q) = \bar{c}(q)] \big)^2$ \Comment{squared agreement weight}
\EndFor
\State \Return $\arg\max_{c \in \mathcal{R}_q} \sum_{e} w_e\, z_e(q,c)$ for each $q \in Q$ \Comment{fused match}
\end{algorithmic}
\end{algorithm}

Because all of the models that I implement for EnsembleLink are open-weight, the algorithm runs locally without API calls and is deterministic and reproducible for a given model version. I use the same default configuration on each benchmark below without task-specific tuning. Specifically, I use the Harrier-OSS-v1 embedding and the Jina v2 and BGE v2-m3 reranker models, and show robustness to alternative models (Appendix~\ref{app:robustness}). The cost of embedding the target data is linear in the dataset size and is incurred only once. The cost of exact retrieval is proportional to the number of queries times the number of records, and the cost of scoring depends only on the number of queries, since each query has at most $2k$ candidates. Linking 1,000 queries to one million records takes about ten minutes on a single GPU (Appendix~\ref{app:scaling}).

\subsection{Benchmarks and Results}

I compare EnsembleLink with fastLink \citep{enamorado2019using}, Splink \citep{linacre2022splink}, and fuzzylink \citep{ornstein2025fuzzylink} on five applications that span diverse linkage domains (Table~\ref{tab:benchmark_examples}). The first links North Carolina voters whose address changed between the 2023 and 2025 voter files to the records of one million voters in the 2025 file, using only name and birth year with a stable voter identifier serving as ground truth. A second version of this task removes the true match for one of every five queries, penalizing links made for queries without a true match. The second application draws on the MaDI benchmark \citep{steiner2026madi}. MaDI provides candidate pairs of records from five different domains, such as company names and products, both as a cleaned and a noisier version. I link each domain separately and average over domains. For the products domain, I also link each complete product table to each of the other tables, requiring each method to find the corresponding match among all records.

\begin{table}[!t]
\centering
\includegraphics{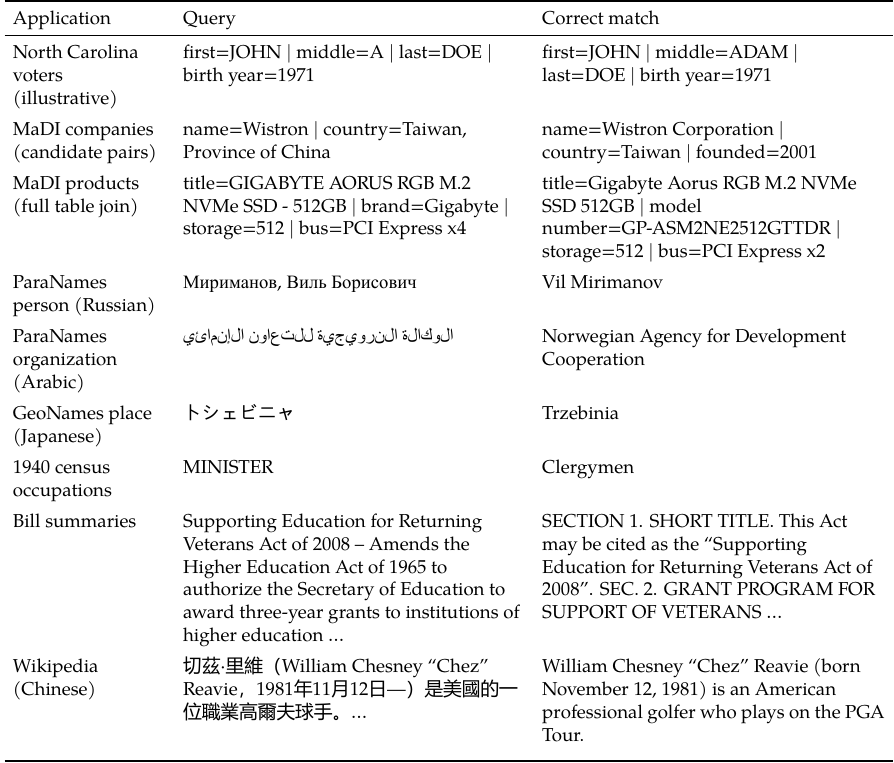}
\caption{Examples of a query and its correct match from each application. The North Carolina row is illustrative and shows the format of the voter records. Every other row comes from the test data.}
\label{tab:benchmark_examples}
\end{table}

The third application draws on ParaNames \citep{saleva2024paranames} and GeoNames \citep{geonames2026} to link the names of people, organizations, and places written in Arabic, Chinese, Hindi, Japanese, Russian, or Spanish to their English forms. In the fourth application, I link the raw occupations transcribed by 1940 census enumerators in the United States to the Census Bureau's 1950 occupational classification (OCC1950), using the codes manually assigned by IPUMS as ground truth. The fifth application links paragraphs of text: first, summaries of congressional bills to the text of the bills themselves \citep{kornilova2019billsum}; and second, Wikipedia articles in the same six languages from the third application to their English counterparts. 

For fairness of comparison, each method is provided the same join fields. I run fastLink and Splink with their package defaults, and fuzzylink with its own pipeline using GPT-5.2, the default model. Running fuzzylink on all benchmarks cost about \$245 in OpenAI API calls, whereas the other methods ran without costs or training labels. Because the active learning of fuzzylink involves using the language model to label candidate pairs drawn from the data being linked, including the test data, I report fuzzylink in two ways: as designed, labeling the test data directly, and trained on the validation data and then applied unchanged to the test data, as the other methods are. I treat the latter as the fairer comparison but report both in full.

Each method assigns to each potential match a score or probability, and a match is accepted only if that value exceeds a predetermined cutoff. For every method except fuzzylink, which has its own internal decision rule, I split each benchmark into a validation set, used to choose each method's cutoff (the cutoff that maximizes F1), and a separate test set, which is used to report results. Choosing the cutoff in this way requires a small labeled validation set. With a fixed cutoff that requires no labels instead, linking a record whenever the rerankers' probability for the top candidate is at least 0.5, EnsembleLink loses only 1.5 F1 points on average. It outperforms fastLink and Splink with their own label-free cutoffs on every benchmark except the North Carolina voters with unmatched records, where these methods retain their advantage (Appendix Table~\ref{tab:common_threshold}). In Appendix~\ref{app:suite}, I document the data, fields, and configurations, as well as precision and recall for every method where defined. 

\begin{figure}[!t]
\centering
\includegraphics[width=\textwidth]{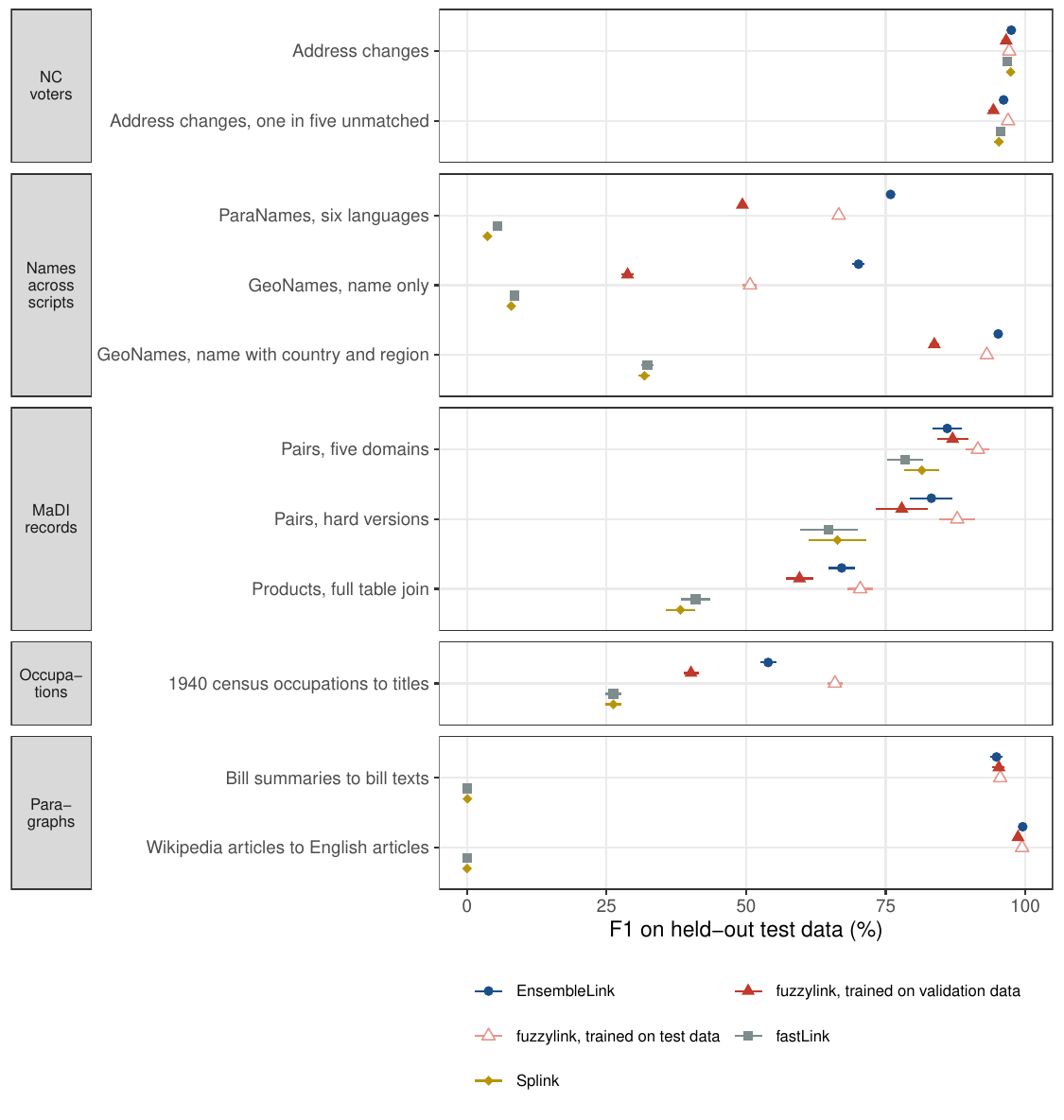}
\caption{Record linkage F1 (\%) on held-out test data, with 95\% bootstrap intervals over queries. Raw values are provided in Appendix Table~\ref{tab:linkage_summary}.}
\label{fig:linkage_summary}
\end{figure}

The results are provided in Figure~\ref{fig:linkage_summary} by benchmark and method, as the F1 score on the held-out test data. For the North Carolina voters task, where records usually agree on name and birth year, every method links nearly every voter correctly. Existing approaches, including those building on the \citet{fellegi1969theory} framework, thus perform well when fields are standardized and rarely missing. When people and place names are written in different languages across sources, fastLink and Splink find few of the correct matches, because there is little shared text between the query and the target records. EnsembleLink finds many of them, and outperforms both versions of fuzzylink across every group of name tasks. In Appendix~\ref{app:errors}, I separate the remaining errors into true matches missed by retrieval and those that made it into the candidate list but were ranked below a wrong candidate in the final stage.

For the MaDI records, which combine a name or title with descriptive fields such as a company's country and founding year, a paper's authors and journal, or a product's brand, model number, and specifications, fuzzylink achieves the highest accuracy when its language model labels the test pairs directly. When fuzzylink is instead trained exclusively on the validation data, EnsembleLink performs comparably on the candidate pairs and better on their noisier versions and on the full join of product tables, without labeled training data. Both methods outperform fastLink and Splink, particularly in the noisier versions of each task.

For the census occupations, where many different spellings correspond to a single standardized occupational title, fuzzylink achieves the highest accuracy when its language model labels the test pairs. When fuzzylink is instead trained only on the validation data, EnsembleLink outperforms it by a wide margin (Appendix~\ref{app:suite}). Both in turn far outperform fastLink and Splink. For the paragraph applications, EnsembleLink and fuzzylink perform comparably, whereas fastLink and Splink find almost none of the matches. With text at this length, string similarity scores nearly every pair as equally dissimilar. 

Because the applications mix several kinds of difference, I also conduct a controlled simulation. In Appendix~\ref{app:simulation}, I link North Carolina voter records and English organization names to systematically altered versions of themselves, applying one kind of alteration at a time. String comparisons work well for spelling errors, nicknames, dropped middle names, and missing values, where the Fellegi--Sunter methods perform comparably to EnsembleLink. However, these string-based approaches fail for more complicated differences, such as when first and last names are swapped between fields (first name ``DOE'', last name ``JOHN''), when a full name appears in a single field with the surname first (``DOE, JOHN ADAM''), and when names are written in different scripts. Abbreviating the words in organization names also costs fastLink and Splink far more accuracy than EnsembleLink. Additional fields help every method, and with the full record, the methods converge.

\section{Ethnicity Inference}

\subsection{Method}

I next turn to the task of ethnicity inference from names. For the main validation, I draw stratified samples of 10,000 voters (2,500 in each of four racial categories: White, Black, Hispanic, and Asian) from the Florida and North Carolina voter files, which record self-reported race at registration and are the most common benchmark for surname geocoding. I draw additional samples from Alabama and South Carolina for supplementary analyses.

For each name, I prompt an LLM to classify it into a constrained set of locally appropriate ethnic groups, such as the race categories used in the U.S. voter files, religious sects in Lebanon, or caste and ethnic groups elsewhere (I list every prompt and compare six templates in Appendix~\ref{app:prompts}). With the default prompt, each request provides the full name and its geographic unit (county in the United States, region or province elsewhere), assigns the model the persona of a domain expert, provides a list of potential categories to select from, and asks for a single classification (Figure~\ref{fig:prompt_us}). I set model temperature to 0. 

\begin{figure}[!t]
\centering
\begin{minipage}{0.95\textwidth}
\small
\begin{verbatim}
You are an expert in American onomastics. Classify the race/ethnicity
of this person based on their full name and county of residence.

Full Name: {name}
County: {county}, {state}

Return ONLY one of these categories: WHITE, BLACK, HISPANIC, ASIAN, OTHER
Do not explain. Just return the single category.
\end{verbatim}
\end{minipage}
\caption{The prompt used for the U.S.\ voter files. The prompts for other countries and categories follow the same structure (Appendix~\ref{app:prompts}).}
\label{fig:prompt_us}
\end{figure}

I compare the results against standard BISG as implemented with the \texttt{wru} R package \citep{khanna2024wru} (surname and county demographics from the 2020 Census), BISG supplemented with first and middle names, which provides the fairest comparison for LLM prompts that use full names, and fBISG \citep{imai2022addressing}. For each method, I assign the classification with the highest probability.

One potential concern is that these voter file records appear in the models' training data, allowing the models to simply recall a voter's race. The advantage over BISG holds, however, on the Alabama and South Carolina voter files, which are never posted publicly. LLM accuracy also holds for voters who registered after the models' training cutoff and for synthetically generated names that do not appear in any voter file (Appendices~\ref{app:access}--\ref{app:rarity}).

\subsection{U.S. Validation}

I report the results for Florida and North Carolina across different information settings (full name, surname, with and without geography), proprietary LLMs, and open-weight LLMs in Table~\ref{tab:ethnicity_accuracy}. With full names and geography under the default prompt, Gemini~3.0 Flash classifies the race of 83.8\% of voters correctly in Florida and 83.5\% in North Carolina, compared to 78.0\% and 78.4\% for name-supplemented BISG.\footnote{A paired McNemar test rejects equal accuracy against both standard and name-supplemented BISG in each state, at $p<10^{-40}$ or lower. Statistics are provided in Appendix~\ref{app:flnc}.} Without first or middle names, the advantage of the LLM over BISG is larger. Many LLMs lose little accuracy when prompted without geographic information, and Gemini does well on surnames missing from the Census tables. The advantage of the approach is therefore greatest in information-constrained settings.

\begin{table}[!t]
\centering
\resizebox{\textwidth}{!}{
\begin{tabular}{lcccccccccccccc}
\toprule
 & \multicolumn{7}{c}{Florida} & \multicolumn{7}{c}{North Carolina} \\
\cmidrule(lr){2-8}\cmidrule(lr){9-15}
Method & Gem. & 4o & 4.1-m & DS4 & GLM5 & Q3.5 & BISG & Gem. & 4o & 4.1-m & DS4 & GLM5 & Q3.5 & BISG \\
\midrule
\multicolumn{15}{l}{\textit{Panel A: Accuracy by input configuration}} \\
Full name + geo & \textbf{83.8} & 78.5 & 70.5 & 76.6 & 74.0 & 74.9 & 78.0 & \textbf{83.5} & 80.5 & 74.1 & 80.0 & 76.8 & 78.3 & 78.4 \\
Full name only & \textbf{82.8} & 78.2 & 69.7 & 77.0 & 73.8 & 75.0 & --- & \textbf{82.3} & 79.6 & 72.9 & 78.5 & 75.8 & 76.8 & --- \\
Surname + geo & \textbf{76.0} & 69.9 & 65.9 & 70.4 & 65.2 & 66.6 & 68.2 & \textbf{74.3} & 69.1 & 66.9 & 71.3 & 66.1 & 67.9 & 68.9 \\
Surname only & \textbf{69.7} & 64.7 & 58.4 & \textbf{69.7} & 62.5 & 62.9 & --- & 68.3 & 64.6 & 60.6 & \textbf{69.1} & 63.5 & 64.2 & --- \\
\midrule
\multicolumn{15}{l}{\textit{Panel B: Recall by race, full name + geo}} \\
White & 88.4 & 88.5 & 90.6 & 86.8 & 90.4 & 86.8 & \textbf{91.5} & 84.4 & 88.2 & 90.5 & 82.4 & 89.2 & 87.3 & \textbf{92.9} \\
Black & \textbf{81.7} & 72.2 & 48.9 & 62.8 & 63.0 & 63.0 & 67.0 & \textbf{80.0} & 68.3 & 52.1 & 71.4 & 63.3 & 64.1 & 65.2 \\
Hispanic & 92.3 & 91.6 & 91.4 & 92.2 & 90.0 & 92.6 & \textbf{93.1} & 84.9 & 85.7 & 83.8 & 84.7 & 81.5 & \textbf{85.8} & 81.2 \\
Asian & \textbf{72.8} & 61.6 & 51.2 & 64.4 & 52.7 & 57.2 & 60.4 & \textbf{84.5} & 79.6 & 70.2 & 81.6 & 73.4 & 76.2 & 74.4 \\
\bottomrule
\end{tabular}}
\caption{Panel A: overall accuracy (\%) by input configuration. Panel B: recall (\%) by racial category with full name and geography, where the BISG column is name-supplemented BISG. Gem.\ = Gemini~3.0 Flash; 4o = GPT-4o; 4.1-m = GPT-4.1 mini; DS4 = DeepSeek-V4; GLM5 = GLM-5-Turbo; Q3.5 = Qwen3.5-Plus. BISG (\texttt{wru} v3.0.3) requires geography and has no counterpart in the name-only rows. Bold marks the best value in each row and state. fBISG and recall for standard BISG are in Appendix~\ref{app:flnc}.}
\label{tab:ethnicity_accuracy}
\end{table}

The largest differences in performance between LLM and BISG arise for Black and Asian voters (Table~\ref{tab:ethnicity_accuracy}, Panel B). In both states, Gemini correctly identifies about 15 percentage points more Black voters than name-supplemented BISG, and 10 to 12 percentage points more Asian voters. Name-supplemented BISG identifies more White voters, and the two methods are close for Hispanic voters. The advantage of LLMs over standard BISG also holds for precision and F1 (Appendix Table~\ref{tab:ethnicity_prf}).

Although the strongest results come from a proprietary model, Gemini~3.0 Flash, there are a number of reasons to prefer open-weight models. These include replicability, by eliminating dependency on models that may not be offered in the future; cost, where open-weight models can run for free conditional on the requisite hardware; and privacy concerns, where researchers may wish to conduct ethnicity inference locally on-device without uploading data to external providers. I therefore run the same classification tasks across a panel of thirty-five models, using samples of 2,000 voters from each of Florida, North Carolina, Alabama, and South Carolina. I use both proprietary and open-weight models, models from different generations of each model family, and models with parameter counts ranging from a few billion to more than a trillion (Table~\ref{tab:durability}).

Of these thirty-five models, thirty, including eleven open-weight ones, outperform standard BISG in all four states. Compared to name-supplemented BISG, the fairer comparison given that each model is prompted with full names, only Gemini 3.8 Flash performs better in all four states at its lowest reasoning setting. With reasoning enabled where the model supports it, three Gemini Flash models do (Appendix Table~\ref{tab:reasoning}). Five other models, including two open-weight models, DeepSeek-V4.1-Flash and DeepSeek-V4-Pro, perform better in three states and fall short by less than one percentage point in the fourth. The two DeepSeek-V4 models therefore make the approach reproducible while also performing about as well as name-supplemented BISG, provided that reasoning is enabled. None of the models with fewer than 100 billion parameters outperform name-supplemented BISG in any of the four states.

\begin{table}[!t]
\centering
\footnotesize
\begin{tabular}{lcccccc}
\toprule
Model & FL & NC & AL & SC & $>$BISG & $>$+names \\
\midrule
\textit{BISG (surname + geography)} & 67.4 & 69.5 & 64.6 & 62.8 \\
\textit{BISG (+ first/middle names)} & 77.8 & 79.9 & 75.2 & 70.5 \\
\midrule
\multicolumn{7}{l}{\textit{Proprietary (API-only)}} \\
Gemini 3.8 Flash & \textbf{81.2} & \textbf{80.3} & \textbf{76.0} & \textbf{75.0} & \checkmark~(4) & \checkmark~(4) \\
Claude Opus 4.8 & 80.7 & 79.7 & 75.6 & 74.6 & \checkmark~(4) & \checkmark~(3) \\
Claude Sonnet 4.6 & 80.5 & 79.8 & 74.6 & 72.5 & \checkmark~(4) & \checkmark~(2) \\
Gemini 3.5 Flash & 80.3 & 80.0 & 72.2 & 72.7 & \checkmark~(4) & \checkmark~(3) \\
Muse Spark 1.3 & 80.3 & 78.8 & 74.5 & 73.1 & \checkmark~(4) & \checkmark~(2) \\
Gemini~3.0 Flash & 80.2 & 80.0 & 74.6 & 73.6 & \checkmark~(4) & \checkmark~(3) \\
GPT-5.5 & 79.4 & 78.1 & 74.9 & 72.3 & \checkmark~(4) & \checkmark~(2) \\
GPT-5.4 & 79.3 & 79.0 & 74.7 & 72.3 & \checkmark~(4) & \checkmark~(2) \\
GPT-4o & 78.7 & 77.5 & 72.6 & 72.1 & \checkmark~(4) & \checkmark~(2) \\
GPT-6 Luna & 78.5 & 76.4 & 71.9 & 70.5 & \checkmark~(4) & \checkmark~(1) \\
Qwen3.7-Plus & 77.4 & 77.2 & 72.8 & 71.8 & \checkmark~(4) & \checkmark~(1) \\
Qwen3.5-Plus & 77.0 & 77.6 & 73.0 & 71.4 & \checkmark~(4) & \checkmark~(1) \\
GPT-5.6 Luna & 76.6 & 75.9 & 70.8 & 70.0 & \checkmark~(4) &  \\
Gemini 3.1 Flash Lite & 76.0 & 75.6 & 73.0 & 70.1 & \checkmark~(4) &  \\
GPT-4.1 mini & 75.3 & 76.8 & 70.5 & 69.5 & \checkmark~(4) &  \\
GPT-5.4 mini & 75.2 & 76.4 & 71.9 & 69.5 & \checkmark~(4) &  \\
Qwen3.8-Flash & 74.9 & 74.8 & 69.9 & 69.2 & \checkmark~(4) &  \\
GLM-5-Turbo & 73.2 & 75.4 & 71.9 & 69.8 & \checkmark~(4) &  \\
Qwen3.5-Flash & 70.0 & 71.8 & 67.6 & 65.8 & \checkmark~(4) &  \\
Muse Spark 1.2 & 43.5 & 45.7 & 41.9 & 39.0 &  &  \\
\hdashline
\multicolumn{7}{l}{\textit{Open-weight}} \\
DeepSeek-V4-Pro (0813) & 78.6 & 78.9 & 73.4 & 72.0 & \checkmark~(4) & \checkmark~(2) \\
DeepSeek-V4 & 78.4 & 79.0 & 73.0 & 72.2 & \checkmark~(4) & \checkmark~(2) \\
DeepSeek-V4.1-Flash & 78.4 & 77.8 & 73.3 & 70.3 & \checkmark~(4) & \checkmark~(1) \\
GLM-5.3 & 77.5 & 76.4 & 72.7 & 72.7 & \checkmark~(4) & \checkmark~(1) \\
Qwen3.5-397B & 77.3 & 78.5 & 72.4 & 69.5 & \checkmark~(4) &  \\
DeepSeek-V3.2 & 75.8 & 76.8 & 70.8 & 70.3 & \checkmark~(4) &  \\
Llama-4 Maverick & 72.6 & 74.5 & 67.3 & 66.3 & \checkmark~(4) &  \\
Qwen3.5-27B & 72.4 & 75.5 & 69.5 & 67.8 & \checkmark~(4) &  \\
Qwen3.8-27B & 71.4 & 73.0 & 67.2 & 66.1 & \checkmark~(4) &  \\
Qwen3.6-35B & 70.9 & 73.4 & 67.7 & 67.7 & \checkmark~(4) &  \\
Qwen3.5-122B & 67.5 & 72.2 & 68.1 & 66.9 & \checkmark~(4) &  \\
Qwen3.5-35B & 67.0 & 69.9 & 65.8 & 64.6 & \checkmark~(3) &  \\
Qwen3.8-2.4T & 66.7 & 66.6 & 61.0 & 59.2 &  &  \\
Qwen3.5-9B & 61.2 & 62.6 & 60.7 & 57.5 &  &  \\
Muse Glimmer 30B & 57.6 & 60.1 & 55.4 & 55.0 &  &  \\
\bottomrule
\end{tabular}
\caption{Accuracy (\%) classifying race from full name and geography, 2{,}000 voters per state. Each model is run at the lowest reasoning setting accepted by the respective API (Appendix~\ref{app:panel_models}). Bold marks the best model in each column. A check mark indicates that the model outperforms standard BISG ($>$BISG) or name-supplemented BISG ($>$+names) in at least one state, with the number of states in parentheses. Responses outside the four categories count as incorrect. BISG rows use \texttt{wru} v3.0.3, without and with first- and middle-name supplements, on the same voters. Model identifiers, reasoning settings, and licenses are in Appendix~\ref{app:panel_models}.}
\label{tab:durability}
\end{table}

More generally capable models are not always more accurate on this task, and more reasoning similarly does not always help. With high reasoning, for instance, several OpenAI models tend to over-predict White, the population majority: GPT-5.4 mini labels 91\% of the balanced sample as White and classifies only 33\% of voters correctly, compared to 73\% with reasoning switched off. Conversely, high reasoning raises accuracy by about two points for Gemini~3.0 Flash and GPT-5.5 (Appendix Table~\ref{tab:reasoning}). Nor do newer releases reliably improve on older ones, with GPT-5.6 Luna and GPT-6 Luna scoring below the older GPT-4o model in every state. Researchers may therefore benchmark candidate models on a small labeled sample before applying the method.

On the names of U.S.\ elected officials with hand-labeled races from \citet{lee2025measuring}, Gemini~3.0 Flash with geography reaches 90.4\% accuracy, higher than BISG (84.5\%), fBISG (85.4\%), and the model proposed by the authors that combines names with facial images (88.8\%) (Appendix Table~\ref{tab:leevelez}).

\subsection{International Validation}

I begin the international validation of this approach with the Lebanese voter file ($n=3{,}676{,}322$), which offers a rare setting outside of the United States with individual-level names and ground truth ethnic identification. The Lebanese voter file records each registered voter's name, locality of residence, and religious sect, and naming patterns are closely connected to sect for many groups. Moreover, the file is not readily available online.\footnote{BISG cannot be fairly benchmarked here, since out-of-sample name-sect frequency tables do not exist. An in-sample benchmarking of BISG is provided in Appendix~\ref{app:lebanon}.} On a stratified sample of 500 voters each from the seven largest sects (covering 96.6\% of all voters), Gemini~3.0 Flash classifies 66.5\% of voters correctly, compared to 44.6--55.5\% for the other models (Appendix Table~\ref{tab:lebanon}). As expected, Gemini classifies voters more accurately where names are more distinctive to a sect (97.2\% of Armenian Orthodox voters) and less accurately where naming conventions have converged between sects (34.2\% of Greek Catholics).

Reserved constituencies in India offer another comparative validation setting with individual-level ground truth. For parliamentary and state assembly elections, members elected from seats reserved for Scheduled Castes (SC, historically disadvantaged Dalit communities) or Scheduled Tribes (ST, Indigenous Adivasi communities) must by law belong to the respective group. For the 131 members elected from these seats in 2024, Gemini~3.0 Flash distinguishes SC from ST with 99.2\% accuracy from names alone, although the other models trail behind (Appendix Table~\ref{tab:india_scst}).

Distinguishing all four caste reservation categories is harder: Gemini~3.0 Flash reaches 75.3\% accuracy on land records from Bihar, where names often carry caste markers, but 57.7\% among village council heads in Rajasthan, where they often do not (Appendix~\ref{app:india}).

Finally, I validate the LLM approach at the aggregate level where no individual-level ground truth is available, by conducting individual-level inference and then aggregating group shares for comparison with published census figures. I obtain full-count voter rolls for India, Uganda, Nepal, Armenia, Chile, and Costa Rica, and prompt a model with each voter's country, name, region, and a list of the relevant ethnic categories. 

The models perform well across these diverse settings where BISG and related approaches cannot be estimated (Appendix~\ref{app:aggregate}). Gemini~3.0 Flash, for instance, recovers India's religious composition with a mean absolute error of 0.3 points, and detects minorities as small as Armenia's Yezidis (1.2\% of the population; Gemini estimates 1.0\%). The models perform poorly where names are not expected a priori to correspond to specific ethnic groups, such as with Spanish-named Indigenous Costa Ricans. Where naming patterns overlap across groups, models tend to over-predict the largest group.
 
\subsection{Scaling}
\label{subsec:scaling}

For large-scale applications where proprietary models are costly and open-weight models perform insufficiently, researchers can classify a small sample of names using a frontier LLM, fine-tune a small open-weight model on those labels, and then run the small model locally at no marginal cost. With about 1,200 labeled names from Gemini~3.0 Flash, fine-tuning raises the accuracy of Qwen3-0.6B, which has almost no zero-shot ability, from 7.3\% to 75.7\% on U.S.\ race predictions, below the 83.3\% of the teacher model (Appendix~\ref{app:distillation}).

\subsection{Uncertainty and calibration}
\label{subsec:uncertainty}

Downstream estimates of group differences require classification probabilities, but proprietary LLMs rarely provide individual token probabilities. Instead, I train a logistic classifier on embeddings of names using LLM-generated labels on a sample of voters. This classifier approaches the accuracy of BISG in Florida and North Carolina without requiring Census tables or human labels. 

\begin{figure}[!t]
\centering
\includegraphics[width=.98\linewidth]{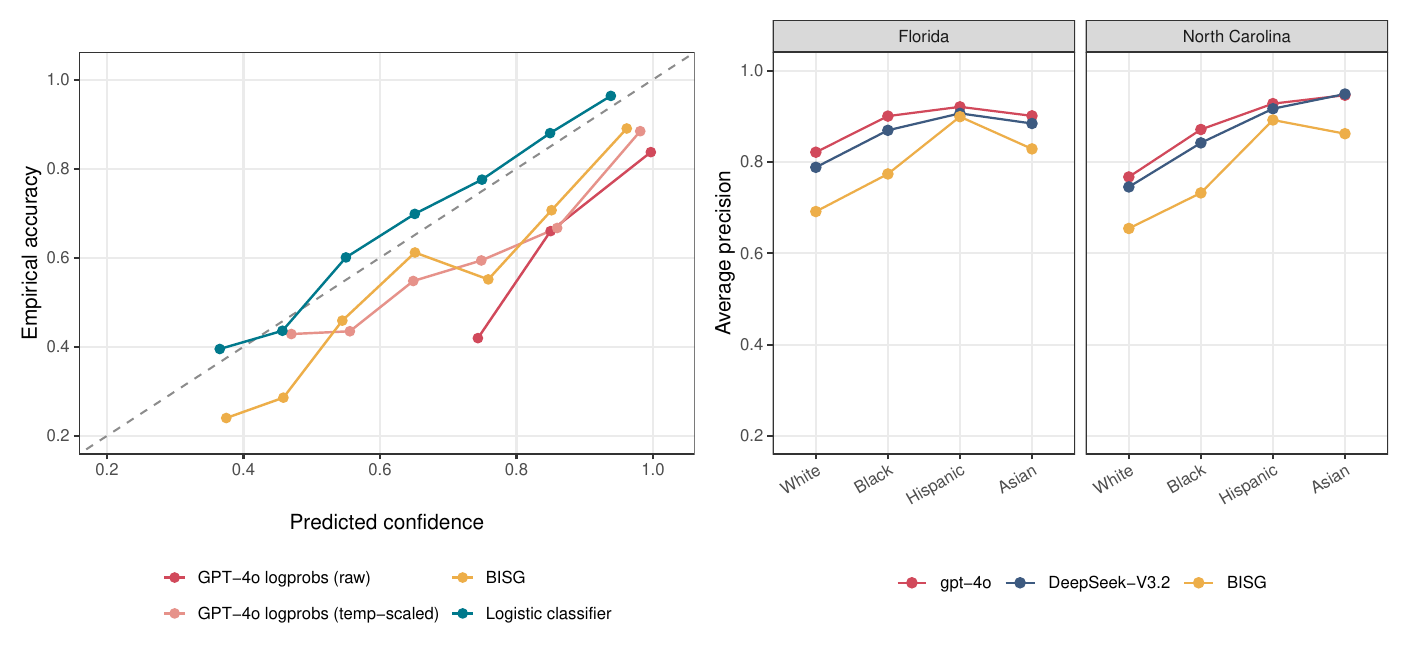}
\caption{Left, probability calibration on the Florida test set, with expected calibration error per method. Right, discrimination in Florida and North Carolina, with one-versus-rest average precision.}
\label{fig:reliability}
\end{figure}

I assess accuracy and calibration (whether a predicted probability of $p$\% is correct $p$\% of the time) separately by state (Figure~\ref{fig:reliability}). In Florida, the raw token probabilities provided by GPT-4o are overconfident with an expected calibration error of 0.16, compared to 0.10 for BISG and 0.03 for the logistic classifier.\footnote{The most accurate proprietary classifier, Gemini, does not provide token log probabilities.} The classifier is similarly well calibrated in North Carolina (Appendix Figure~\ref{fig:reliability_nc}). With respect to discrimination, various proprietary and open-weight models rank group members ahead of non-members at least as well as standard BISG in every category and state (Appendix~\ref{app:probrank}).

Finally, drawing repeatedly from the classifier's probabilities yields multiple-imputation intervals for downstream estimation whose validity is conditional on that classifier. These intervals contain the true racial gap for only five of the twelve turnout and income gaps, and for none of the six party registration gaps (Appendix~\ref{app:mi_intervals}). They do not capture systematic classification error or uncertainty across classifiers, and should not be interpreted as generally valid confidence intervals. 

\subsection{Misclassification bias and downstream estimates}

\citet{argyle2024misclassification} show that BISG misclassification is tied to neighborhood socioeconomic status, such that minorities in wealthier and more educated areas are disproportionately classified as White. When these predicted probabilities are used in estimating downstream quantities, this error biases the estimates. Because BISG relies on a Bayesian prior defined by the racial composition of each voter's geographic unit, these errors are mechanically tied to geography. LLM-based inference can optionally use geography as an input but does not force any such prior, and should therefore reduce this sort of bias. 

I first replicate the known bias for BISG. Among Black and Hispanic voters, BISG labels voters as White more often in higher-income and more educated neighborhoods, at both the ZCTA and census tract level (Appendix~\ref{app:gradient}). Whether LLM-based inference reduces this relationship depends on the group. The reduction from using LLMs instead of BISG is largest for Black voters, where the relationship is nearly halved across North Carolina, Alabama, and South Carolina. This happens because Black surnames overlap most with those of White voters, requiring BISG to rely the most on geography. The LLM does not reduce the relationship for Hispanic voters, where Hispanic voters with White-sounding names are themselves concentrated in wealthier neighborhoods, and the reduction is negligible for Asian voters, whose names are sufficiently distinctive and recognizable for BISG not to rely on geography. 

\begin{figure}[!t]
\centering
\includegraphics[width=\linewidth]{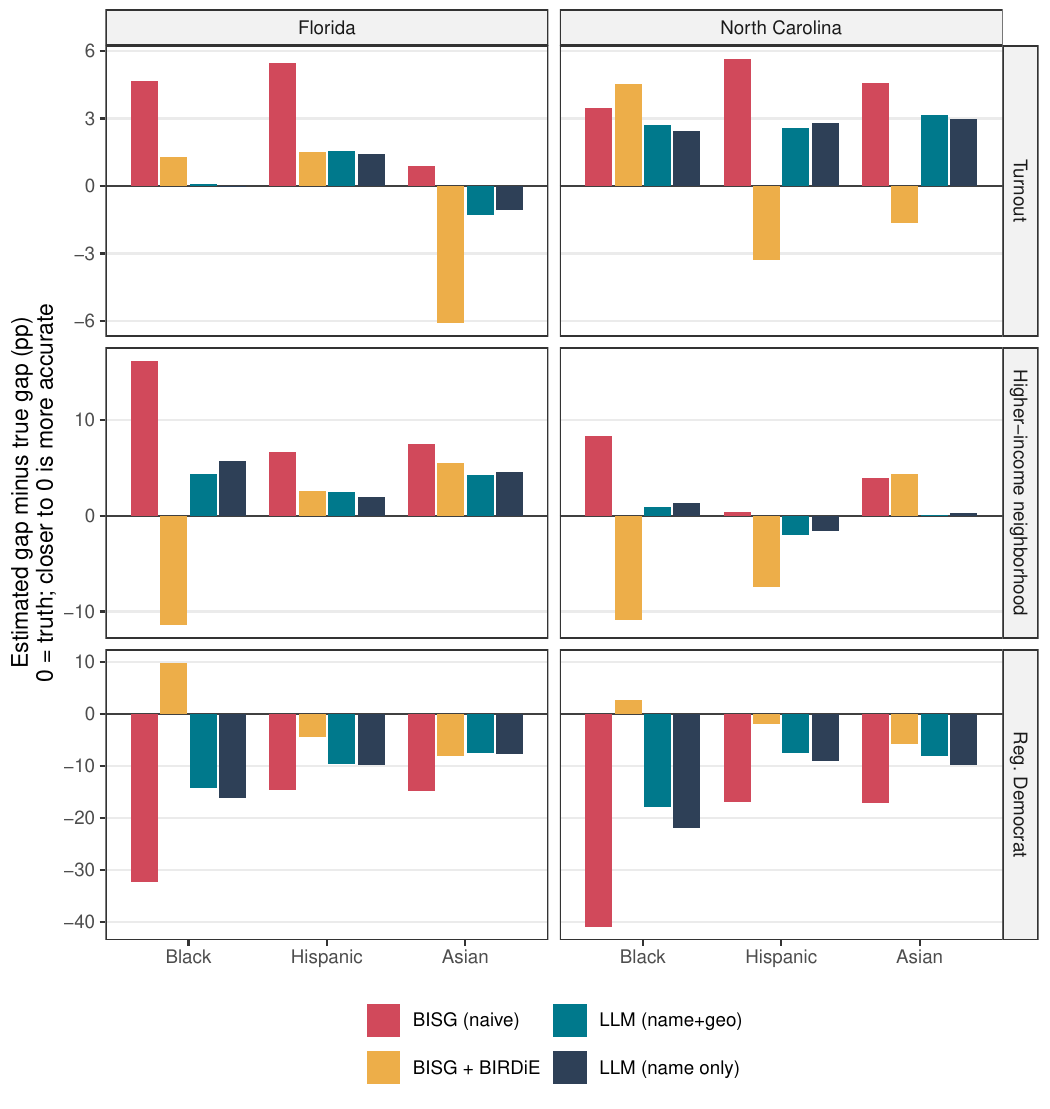}
\caption{Estimated gap between each minority group and White voters minus the gap measured with true race. Frequent turnout and living in an above-median-income ZCTA are the geographic outcomes, and being a registered Democrat is the non-geographic outcome (Appendix~\ref{app:birdie_llm}).}
\label{fig:downstream}
\end{figure}

These gradients bias downstream estimates of racial gaps in different outcomes (Figure~\ref{fig:downstream}). I consider two outcomes whose gaps between each group and White voters operate through geography (turnout and living in a higher-income neighborhood) and one whose gap does not (being a registered Democrat). I compare the gap estimated from the racial predictions of each method to the gap estimated with the ground truth race. 

I find that BISG erroneously attenuates the racial gaps in these outcomes toward zero, as documented by \citet{argyle2024misclassification}, because systematic misclassification moves different groups toward one another. On the geographic outcomes, the LLM classifications produce racial gaps closest to the truth. BISG with BIRDiE \citep{mccartan2025birdie}, the standard correction for BISG, overshoots in the opposite direction. BIRDiE as applied here, without geography in the outcome model, assumes that the outcome is independent of name and geography given race. This assumption fails for outcomes that vary geographically. 

On party registration, by contrast, BIRDiE is the most accurate, having been designed for this kind of outcome, although the LLM still outperforms BISG alone by a considerable margin for Black voters and by a smaller margin for Hispanic and Asian voters. When applied to the probabilities of a name-only logistic classifier trained on LLM-generated labels, BIRDiE recovers the racial party gap more accurately than any other method (Appendix~\ref{app:birdie_llm}). The gradient and geographic-gap results hold across a number of different LLMs (Appendix Table~\ref{tab:gradient_models}).

\section{Agentic AI for Validation}

Validating the outputs of either record linkage or ethnicity inference from names typically requires hand-coding a random sample of observations. As an alternative, I suggest an agentic workflow in which an LLM agent with web search enabled (here Claude Code) checks a sample of outputs one by one, recording a verdict on the accuracy as well as its reasoning and the sources that it consulted. 

For record linkage, I asked the agent to check whether each match is correct based on external information, that is, whether a name in its original script and the English name to which EnsembleLink matched it refer to the same entity. I provided the agent with 100 matches from the held-out ParaNames test set, half of them from the quarter in which EnsembleLink is least confident. The agent did not reach a judgment for about a third of the pairs, most of them obscure entities or names shared by several people or places. Where the agent did reach a judgment, it judged correctly (based on the ground truth) in more than nine cases out of ten (Appendix Table~\ref{tab:agentic}). Two independent runs returned the same judgment for nearly nine pairs in ten.

For the task of ethnicity inference, I asked the agent to check Gemini's racial classifications of 108 local officials from \citet{lee2025measuring} against web sources. The agent found an externally reported race for 105 of them, of which Gemini's classification matched exactly for 103. The agent also flagged four elected officials whose ``ground truth'' labels in the original dataset proved to be wrong. The workflow can thus also serve as an audit of the ground truth data. 

While agentic validation does not fully replace human judgment, it is repeatable, capable of generating detailed documentation, and costs at most a few dollars for a sample of the present size. In the replication materials, I include the samples, prompts, scripts, and output reasoning and documentation for each case. 

\section{Discussion}

I have demonstrated that pre-trained language models can perform two common data tasks in social science, record linkage and ethnicity inference from names, without requiring task-specific training data. Their accuracies at these problems across diverse contexts approach or exceed those of methods requiring labeled examples or name tables. The results are consistent with a ``world knowledge'' hypothesis that I advance in Section~\ref{sec:why}. I release R and Python packages implementing EnsembleLink for record linkage, as well as replication data and code for both tasks. I also release the fine-tuned Qwen3 and logistic classifier models for racial inference in the U.S.

Choosing models involves a trade-off between accuracy and reproducibility, in particular for ethnicity inference. However, open-weight models allow for local replication and increasingly approach the performance of proprietary models across domains and contexts. For record linkage with EnsembleLink, all models implemented and benchmarked here are open-weight.

For the task of ethnicity inference, accuracy depends on how strongly names encode the group in question. Researchers should validate specific models before applying the method at scale, using hand-coded samples, the agentic validation approach proposed above, or agreement across models (Appendix~\ref{app:agreement}). LLMs may also encode societal biases in the association of names with groups. This might disadvantage individuals with atypical names, such as multiracial or culturally assimilated individuals.

Both record linkage and ethnicity inference tend to involve sensitive personal information, and the methods differ in how they handle this. EnsembleLink, fastLink, and Splink all run entirely on the researcher's own machine. By contrast, fuzzylink sends the text of candidate pairs to a commercial model provider to label training data, as does the ethnicity analysis when it uses proprietary models. Researchers using either approach should confirm that their data agreements permit such data transfers. Open-weight models and local distilled classifiers avoid such concerns, at some cost in accuracy for the ethnicity inference task. 

For ethnicity inference, BISG remains a well-established and generally well-performing option where detailed name and geography data are available, such as in the U.S. with respect to the common racial categories. For record linkage, fastLink and Splink remain strong choices when records share several clean information fields in a common format. EnsembleLink performs better when names are written in different scripts, split differently across fields, or abbreviated. EnsembleLink also performs much better than fastLink and Splink where records are more diverse, such as paragraphs of text, and does not require labels or calls to external models. Using a large language model to label data being linked, as fuzzylink does, is most useful where a match depends on knowledge about the entity that smaller open-weight models may lack, and where data agreements permit sending records to external providers. 


\newpage
\bibliography{references}

\newpage
\begingroup
\pretocmd{\subsection}{\FloatBarrier}{}{}
\pretocmd{\subsubsection}{\FloatBarrier}{}{}

\input{appendix_body}

\endgroup

\end{document}

%% file: header.tex
\usepackage[letterpaper, left=1.25in, right=1.25in, top=1.2in, bottom=1.2in]{geometry}
\usepackage[colorlinks=true, citecolor=blue, linkcolor=blue, urlcolor=blue]{hyperref}
\usepackage[dvipsnames]{xcolor}

\usepackage{natbib} 
\usepackage{caption}
\usepackage{subcaption}

\usepackage{amsmath, amssymb, mathtools, bbm}
\usepackage{amsfonts}
\usepackage{bm} 
\usepackage{physics}  
\usepackage{mathpazo}

\usepackage{amsthm}

\theoremstyle{plain}

\theoremstyle{definition}

\theoremstyle{remark}

\usepackage{multirow} 
\usepackage{float}

\usepackage{tikz}
\usetikzlibrary{shapes, arrows.meta, positioning, calc}

\usepackage{graphicx}
\usepackage{booktabs}
\usepackage{arydshln}
\usepackage{microtype}
\usepackage{setspace}
\usepackage{placeins}
\usepackage{comment}

\makeatletter
\newcommand{\assumplabel}[2]{%
   \protected@write \@auxout {}{\string\newlabel{#1}{{#2}{\thepage}{#2}{#1}{}}}%
   \hypertarget{#1}{#2}%
}
\makeatother

\usepackage{longtable}

%% file: appendix_body.tex
\etocdepthtag.toc{appendix}
\etocsettagdepth{mainpaper}{none}
\etocsettagdepth{appendix}{subsection}
\begin{center}{\Large\bfseries Appendix}\end{center}
\renewcommand{\contentsname}{Contents}
\tableofcontents
\clearpage

\appendix

\section{Record Linkage}
\label{app:linkage}

\subsection{Benchmark Data, Configurations, and Detailed Results}
\label{app:suite}

I report the values plotted in Figure~\ref{fig:linkage_summary} in Table~\ref{tab:linkage_summary}.

\begin{table}[!t]
\centering
\footnotesize
\adjustbox{max width=\textwidth}{\input{tables/tab_linkage_summary.tex}}
\caption{Record-linkage F1 (\%) on held-out test data, as plotted in Figure~\ref{fig:linkage_summary}. Bold indicates the best method in each row. Thresholds are chosen on validation data, except for fuzzylink (test data), which labels test pairs with GPT-5.2 and uses the decision rule of the package. MaDI rows average over domains (pairs) or target tables (full join).}
\label{tab:linkage_summary}
\end{table}

\subsubsection*{North Carolina voters}

Queries are voters whose house number or street name changed between the 2023 and 2025 statewide voter files. I link voters from 2023 on first, middle, and last name, suffix, and birth year to target data of one million records in 2025. The target data contain each voter's own record with the same information, as well as a stable voter identifier that serves as the linkage ground truth. The validation data comprise 1,000 such voters and the test data 2,000. I also include 500 voters in each sample whose address did not change as a diagnostic. Every method links these voters almost perfectly, with an F1 between 97.9 and 99.4 (Table~\ref{tab:nc_detail}). In the second test condition, I replace the target records of 20\% of the queries with records of other voters, such that those queries have no correct match.

\begin{table}[!t]
\centering
\small
\adjustbox{max width=\textwidth}{\input{tables/tab_nc_detail.tex}}
\caption{North Carolina voters, held-out test results (\%). The first two panels are voters whose address changed, and the third panel is the baseline sample of voters whose address did not change. False links are the share of queries without a match in the target data that receive a link. Bold indicates the highest F1 in each panel.}
\label{tab:nc_detail}
\end{table}

\subsubsection*{MaDI}

The five MaDI domains pair records of companies, video games, music albums, scholarly papers, and products from different sources \citep{steiner2026madi}. Each record includes the following information by domain: name, country, city, and founding year for companies; name, platform, developer, and release year for video games; name, artist, release date, and number of tracks for music; title, authors, year, journal, and first and last page for papers; and title, brand, model, model number, product type, and fifteen technical specifications (such as chipset, memory, storage, dimensions, and weight) for products.

I apply each method to each separate domain, using the thresholds determined on validation pairs, separately for the easy, medium, and hard versions of each task. These difficulty levels alter the records with different types of noise. For the full project joins, I link each of 812 source records (321 validation and 491 test queries) against the entirety of each of three target tables. A source record without a match in a table is counted as unmatched.

\begin{table}[!t]
\centering
\small
\adjustbox{max width=\textwidth}{\input{tables/tab_madi_detail.tex}}
\caption{MaDI pair classification, held-out test F1 (\%) by domain. Bold indicates the highest value in each column of each panel.}
\label{tab:madi_detail}
\end{table}

\begin{table}[!t]
\centering
\small
\adjustbox{max width=\textwidth}{\input{tables/tab_products_detail.tex}}
\caption{Full join of the MaDI product tables, held-out test results (\%) by target table. False links are the share of source records without a match in the target table that receive a link (dashes where every source record has a counterpart). Bold indicates the highest F1 in each panel.}
\label{tab:products_detail}
\end{table}

\subsubsection*{ParaNames and GeoNames}

ParaNames (release v2024.05.07.0) contains the names of Wikidata entities in different languages. For each of Arabic, Chinese, Hindi, Japanese, Russian, and Spanish languages and each entity type (person, organization, place), I define the queries as names in that language. The corresponding target data to which the entities are then linked comprise 10,000 English-language names, including names alphabetically adjacent to each true match. The validation and test data are 500 and 1,000 queries long each. GeoNames provides populated places with names in the same six languages. Each GeoNames task is estimated first with the place name alone and second with the place name, country, and first-level region. In the second condition of each task, I ensure that 20\% of queries have no true match.

\begin{table}[!t]
\centering
\small
\adjustbox{max width=\textwidth}{\input{tables/tab_names_detail.tex}}
\caption{Names across languages, held-out test results (\%), averaged over tasks. The transliteration baseline converts each name to Latin letters before the Jaro-Winkler comparison. Bold indicates the highest F1 in each panel.}
\label{tab:names_detail}
\end{table}

\subsubsection*{Census occupations}

For the census occupations task, the queries are the occupations that census enumerators recorded for household heads in the 1940 full count, as transcribed by IPUMS. The target data contain the 269 occupation titles of the 1950 Census Bureau classification (OCC1950), and the ground truth is the code assigned by IPUMS. Excluding blank entries and the codes that record no occupation (keeping house, retired, unemployed, and similar), I draw 2,000 validation and 5,000 test household heads at random. I link with each method using only the occupation name.

\begin{table}[!t]
\centering
\small
\adjustbox{max width=\textwidth}{\input{tables/tab_occupations_detail.tex}}
\caption{1940 census occupations linked to the 1950 occupation titles, held-out test results (\%). Top-1 accepts the highest-scoring title for every query. ``Top-1, near title'' restricts top-1 to queries whose transcribed occupation is a near-exact string match to an OCC1950 title, and ``Top-1, other'' to the remaining queries. Bold indicates the highest F1.}
\label{tab:occupations_detail}
\end{table}

\subsubsection*{Paragraphs}

In the first paragraphs task, the queries are Congressional Research Service summaries of U.S.\ bills from BillSum \citep{kornilova2019billsum}, and the target data contain the opening text of each bill. In the second task, the queries are the opening sections of Wikipedia articles in Arabic, Chinese, Hindi, Japanese, Russian, and Spanish about ParaNames entities, and the target data contain the opening sections of the corresponding English articles. Each task has 500 validation and 1,500 test queries against 5,000 target texts, all reduced to 480 tokens in order that the query and a candidate together fit the input limit of the rerankers. As elsewhere, in the second version of each task, 20\% of queries are missing a corresponding true match. 

\begin{table}[!t]
\centering
\small
\adjustbox{max width=\textwidth}{\input{tables/tab_longform_detail.tex}}
\caption{Paragraphs, held-out test results (\%). Top-1 accepts the highest-scoring text for every query. Bold indicates the highest F1 in each panel.}
\label{tab:longform_detail}
\end{table}

\subsubsection*{Configurations}

fastLink (version 0.6.1) and Splink (version 4.0.12) compare each field separately, using Jaro-Winkler comparisons for the names with cut-points of 0.94 and 0.88 (the fastLink defaults), exact comparison of suffix, and graded comparison of birth year. Parameters are estimated using expectation-maximization and without labels. Lowering the cut-point for full agreement to 0.90, a common choice in applied work, changes the F1 on validation set voters by less than a point for both methods (Table~\ref{tab:jw_sensitivity}). On the North Carolina voters, both methods score the pairs that share at least one of five blocking keys (surname with first initial, first name with surname initial, and combinations of names with birth year), which capture the true record for nearly every query.

I run fastLink with three corrections to the estimation procedure of the current fastLink release. Without them, the algorithm can assign estimated probabilities to the wrong comparison levels when those levels first appear out of order; produce negative expected counts through rounding; and confuse a probability of 1 with an agreement code (probabilities were written into the vector of agreement codes one level at a time; a substituted probability of exactly 1 could then be matched as agreement code 1 on the next pass). Each correction leaves the results unchanged wherever these conditions do not arise. I have proposed the corresponding fixes to the original package's implementation.

I run fuzzylink (version 0.4.1) on its own pipeline, with text-embedding-3-large embeddings and labels generated by GPT-5.2. Across all validation and test runs, the API calls of fuzzylink cost about \$245: \$134 on the name tasks, where each test condition pairs 1,000 queries with 10,000 target records, \$93 on MaDI, \$11 on the paragraph tasks, \$5 on occupations, and \$2 on the North Carolina voters.

I report precision, recall, and F1 in Tables~\ref{tab:nc_detail}, \ref{tab:products_detail}, \ref{tab:names_detail}, \ref{tab:occupations_detail}, and~\ref{tab:longform_detail}, and the share of queries without a match in the target data that receive a link in Tables~\ref{tab:nc_detail}, \ref{tab:products_detail}, and~\ref{tab:names_detail}. I report top-1 accuracy and PR-AUC for every method and benchmark in Table~\ref{tab:linkage_ranking}.

\begin{table}[!t]
\centering
\small
\adjustbox{max width=\textwidth}{\input{tables/tab_linkage_ranking.tex}}
\caption{Top-1 accuracy and PR-AUC (\%) on held-out test data, without a decision threshold. Top-1 is the share of queries with a counterpart whose highest-scoring candidate is the true record. PR-AUC is the average precision of the top-scoring link of each query, ranked by the score that each method thresholds, with queries that have a counterpart as the recall denominator. MaDI pairs have no top-1 because each record has a single labeled candidate. Bold indicates the best method in each row.}
\label{tab:linkage_ranking}
\end{table}

\begin{table}[!t]
\centering
\small
\adjustbox{max width=\textwidth}{\input{tables/tab_jw_sensitivity.tex}}
\caption{Precision, recall, and F1 (\%) of fastLink and Splink by the Jaro-Winkler cut-point for full agreement on names, North Carolina validation voters whose address changed. The cut-point for partial agreement is fixed at 0.88. Bold indicates the highest F1 of each method in each condition.}
\label{tab:jw_sensitivity}
\end{table}

\subsubsection*{Cutoffs without labels}

In the main results, I choose each method's cutoff on the labeled validation data of each benchmark. In Table~\ref{tab:common_threshold}, I report held-out F1 using pre-determined cutoffs. I show performance using a cutoff chosen on the validation data of the other benchmarks only, one cutoff for all benchmarks, a fixed reranker probability of 0.5, and no cutoff at all. I report results for fastLink and Splink using a match probability of 0.5.

\begin{table}[!t]
\centering
\small
\adjustbox{max width=\textwidth}{\input{tables/tab_common_threshold.tex}}
\caption{Held-out F1 (\%) under different cutoff rules. Task-specific cutoffs are chosen on each benchmark's validation data, as in Table~\ref{tab:linkage_summary}, and leave-one-out cutoffs on the other benchmarks' data. Reranker $\geq$ 0.5 does not require any labels. fastLink and Splink link at a match probability of 0.5. Bold indicates the highest F1 in each row, the highest mean, and the smallest losses.}
\label{tab:common_threshold}
\end{table}

The fixed reranker probability lowers the F1 of EnsembleLink by 1.5 points on average. On the North Carolina voters task with unmatched records, EnsembleLink loses about 6.9 points in accuracy, linking together too many records that have no true match. At a match probability of 0.5, fastLink and Splink link almost none of the records for the cross-scripts, occupations, or paragraphs tasks, because their match probabilities rarely reach 0.5 in these domains.

\subsection{Controlled Simulation}
\label{app:simulation}

To separate performance on the kinds of record differences mixed together by the applications, I conduct a controlled simulation. I draw target data of 100,000 North Carolina voter records, use 2,000 of them as queries (500 for validation and 1,500 for testing), and add 500 queries whose records are not in the target data. Each query copies the voter's own record with one kind of difference applied at a set rate and nothing else changed. Changes in accuracy across conditions are then attributable to the kind of transformation applied. 

The different transformations that I apply are: spelling errors (one to three random insertions, deletions, substitutions, or transpositions in each of first and last name); nicknames from a fixed list of standard English forms (Robert to Bob, Margaret to Peggy, etc.); changes to the structure of the name (a first initial, a dropped middle name, first and last names exchanged between fields, a compound surname shortened to the first part); a surname replaced by another drawn from the file; the names written in Cyrillic letters; and values missing at random, with a probability that rises with age, or, for birth year, with a probability that depends on the age itself. A final set of transformations combines moderate amounts of these differences and varies the fields available, from names alone to names, birth year, city, ZIP code, and street.

In two further sets, I alter the form of the record while keeping the content. In the first, the query's names are provided in a single field, surname first after a comma or in natural order, while the target data keep separate fields. I provide the methods that compare different fields the target names joined in natural order as a single field. In the second, I define the target data as 20,000 English organization names from ParaNames, and define each query as the name with standard abbreviations of the words (Association as Assn, University as Univ) or as the acronym of the first letters. I have fastLink compare every query with every record, while I have Splink block on the leading characters of the names and on birth year, except for organization names.

\begin{figure}[!t]
\centering
\includegraphics[width=.95\linewidth]{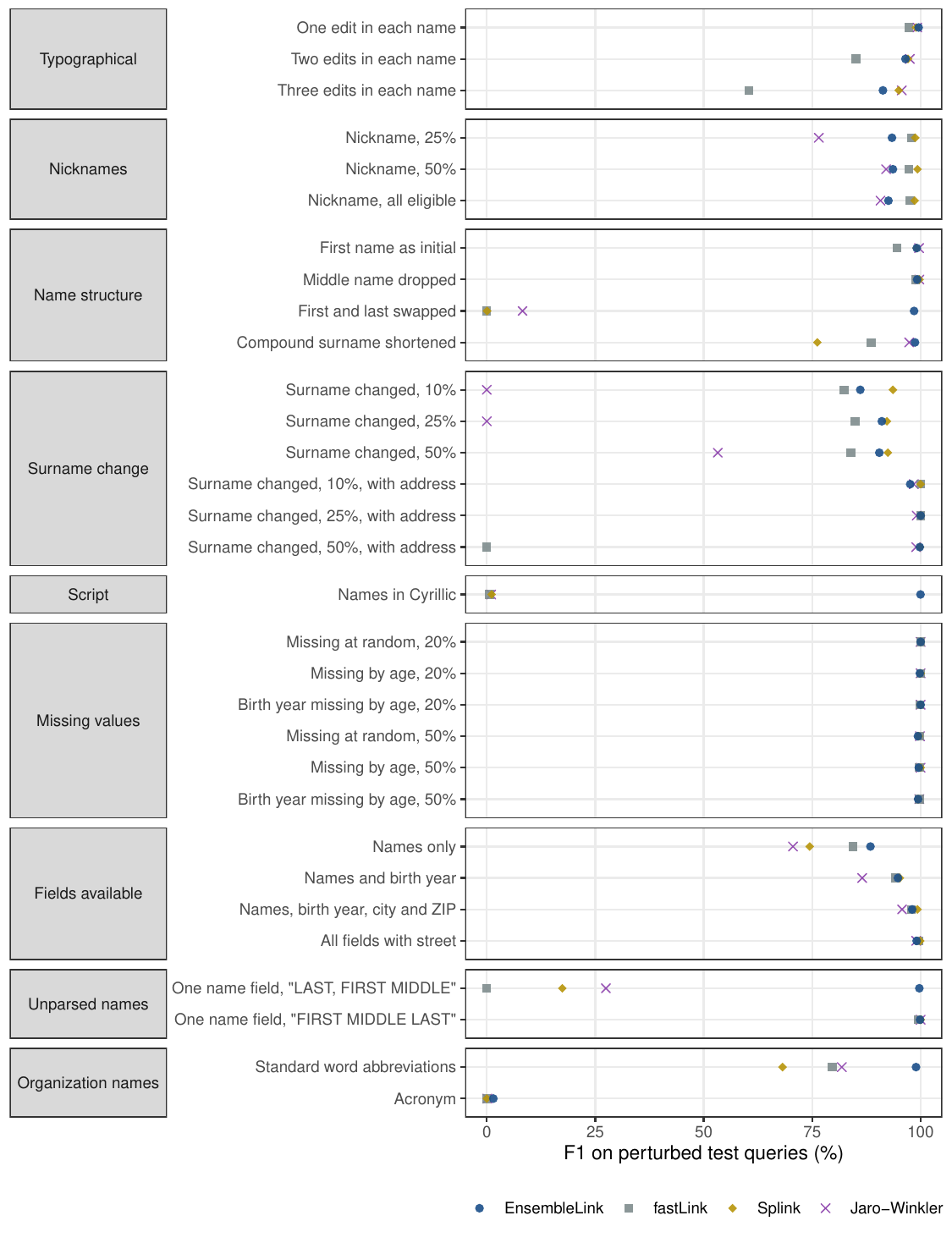}
\caption{F1 on the altered test queries of each simulation condition. Jaro-Winkler averages the similarity of the fields that both records observe. Conditions under ``Fields available'' combine spelling errors, nicknames, and replaced surnames.}
\label{fig:simulation}
\end{figure}

In Figure~\ref{fig:simulation}, I show that the Fellegi-Sunter methods and EnsembleLink perform similarly under conditions of light spelling errors, dropped middle names, and missing values. As the number of spelling errors increases, Splink leads slightly and fastLink declines the fastest. Under nicknames, both Fellegi-Sunter methods lead by about four to six points. However, only EnsembleLink is able to link names that have swapped fields or changed language script. Additional fields of information for linking on raise the F1 of every method, and the methods converge on one another in performance with the full record. 

A replaced surname favors Splink when only names and birth year are available. With half of the surnames replaced and the full record available, fastLink requires every field to agree for a match, and so the model links none of the altered queries. Every method sufficiently links names that are provided in a single field in natural reading order, but only EnsembleLink links the records well where the surname is provided first. EnsembleLink also links abbreviated organization names at a higher F1 score than the other methods. No method consistently links acronyms capably. 

\subsection{Choice of Models}
\label{app:robustness}

On the validation data of every benchmark, I substitute recent open models for each component, holding constant all else, and test different model-based changes to the procedure (Table~\ref{tab:model_swap}). 

Of the model substitutions, adding zerank-2 as a third reranker raises validation F1 the most, especially on the cross-script name linkage. However, zerank-2 has four billion parameters and runs seven to nine times slower than the other rerankers, so it is excluded from the default configuration. Substitutions change the results most on the names and least on the North Carolina voters, where every configuration falls within a point of the default implementation. GLiNER2.5-Decide, a classifier built on an English encoder, loses most of the names when it replaces the rerankers at the second stage. 

\begin{table}[!t]
\centering
\footnotesize
\resizebox{\textwidth}{!}{\input{tables/tab_model_swap.tex}}
\caption{Change in validation F1 (percentage points) from the baseline configuration (Harrier-OSS-v1 embedding; Jina Reranker v2 and BGE Reranker v2-m3 as rerankers; the margin between the top two candidates as match confidence).}
\label{tab:model_swap}
\end{table}

Of the procedural changes tested, only the rank-average confidence adds at least marginal performance gain on some tasks without greatly detracting from performance on others. Specifically, it raises the F1 score most where some queries lack a true match in the target data. Dropping the string-distance expert or the hubness-corrected dense expert, adding a word-level expert for identifiers, checking each match in both directions, and adding a second dense query written in Latin letters all failed to improve on at least one task without costing performance on another (not shown in the table).

\subsection{Ensemble Ablation}
\label{app:linkage_ablation}

I define the default expert set to be strong in different cases. The embedding and cross-encoder scores perform well on matches that depend on semantic meaning or on understanding across languages, but they are sometimes confidently wrong on pairs that differ only marginally with respect to the text itself (e.g., a few typos). In such cases, string similarity is more reliable, while string similarity can be penalized in cases where differences between records are often greater than simple character-level errors. 

\begin{table}[!t]
\centering
\small
\adjustbox{max width=\textwidth}{\input{tables/tab_linkage_ablation.tex}}
\caption{Top-1 accuracy (\%) of each expert and of the fusion on held-out test queries that have a counterpart. Names average over tasks. Bold indicates the highest value in each column.}
\label{tab:linkage_ablation}
\end{table}

I compare top-1 accuracy for each expert alone and for the fusion with equal and with agreement weights (Table~\ref{tab:linkage_ablation}). On the North Carolina voters, the character n-gram expert is the strongest single expert and Jaro-Winkler the weakest, while both character-level experts are weak across scripts. The reranker ensemble ties the dense expert on the voters task and outperforms every other expert across scripts. However, both fusion approaches exceed the best single expert on every task, and the agreement weights match or slightly exceed equal weights. 

\subsection{Retrieval Method Comparison}
\label{app:retrieval}

Dense retrieval using cosine similarity places ``OKC'' near ``Oklahoma City'' in embedding space, but the typo ``Montegomery'' need not necessarily locate near ``Montgomery''. Sparse retrieval on TF-IDF vectors of character n-grams, conversely, easily matches ``Montgomery'' to ``Montegomery'', but cannot match ``OKC'' to ``Oklahoma City'' since the two strings do not share any n-grams.

With $k=30$ candidates from each retriever, the union of the candidate lists comprises up to sixty candidates per query after deduplication. To test whether the sparse retriever is even necessary, I compare this design on the validation data using dense retrieval alone, keeping unchanged every other step in the algorithm (Table~\ref{tab:retrieval_ablation}).

\begin{table}[!t]
\centering
\resizebox{\textwidth}{!}{
\begin{tabular}{l|ccc|ccc|ccc|c}
\toprule
 & \multicolumn{3}{c|}{Retrieval recall} & \multicolumn{3}{c|}{Top-1} & \multicolumn{3}{c|}{F1} & Extra pairs \\
Benchmark & Both & Dense 60 & Dense 30 & Both & Dense 60 & Dense 30 & Both & Dense 60 & Dense 30 & (Dense 60) \\
\midrule
North Carolina voters & 99.8 & 99.6 & 99.5 & 97.5 & 97.5 & 97.4 & 97.0 & 97.0 & 97.0 & +8\% \\
ParaNames & 92.8 & 91.2 & 88.9 & 76.4 & 76.2 & 75.5 & 77.3 & 77.2 & 76.6 & +7\% \\
GeoNames & 95.0 & 93.8 & 92.1 & 82.5 & 81.9 & 80.9 & 83.0 & 82.5 & 81.8 & +20\% \\
Products join & 99.4 & 99.8 & 99.3 & 66.3 & 67.1 & 67.1 & 66.6 & 67.6 & 67.4 & +48\% \\
Occupations & 82.2 & 82.7 & 77.8 & 53.9 & 53.6 & 53.8 & 56.1 & 56.2 & 56.1 & +15\% \\
BillSum & 100.0 & 100.0 & 100.0 & 94.4 & 95.0 & 95.3 & 92.8 & 93.2 & 93.2 & +19\% \\
Wikipedia & 100.0 & 100.0 & 100.0 & 100.0 & 100.0 & 100.0 & 99.6 & 99.7 & 99.8 & +3\% \\
\midrule
Mean & 96.4 & 96.0 & 94.8 & 85.2 & 85.2 & 84.9 & 85.2 & 85.2 & 85.0 & \\
\bottomrule
\end{tabular}}
\caption{Retrieval with the dense and sparse candidates combined (``Both'', the default) and with the dense top 60 or top 30 alone, validation data (\%). ``Retrieval recall'' is the share of queries whose true match enters the candidate list. ``Extra pairs'' is the increase in pairs scored by the rerankers under ``Dense 60'' relative to ``Both''.}
\label{tab:retrieval_ablation}
\end{table}

On average, dense retrieval alone with 60 candidates matches the top-1 accuracy and F1 of the default union approach. However, on the short names of ParaNames and GeoNames, dense retrieval alone misses the true match more often, and on GeoNames names without country or region (pooled with the contextual GeoNames task in the table), top-1 accuracy falls by about a point. With 60 dense candidates, the rerankers also score more pairs, and with 30 candidates, accuracy falls on every name benchmark. The sparse retriever therefore recovers name matches that dense retrieval misses, at a lower cost than retrieving additional dense candidates.

\subsection{Transliteration Ablation}
\label{app:translit}

In the default implementation of EnsembleLink, names are transliterated into Latin letters only for use by the sparse retriever and the two string similarity experts (character n-grams and Jaro-Winkler). Regardless of whether transliteration is anywhere performed, the embedding model and the rerankers are always provided only the original text. 

\begin{table}[!t]
\centering
\small
\begin{adjustbox}{max width=\textwidth}
\begin{tabular}{l|cc|cc|cc}
\toprule
 & \multicolumn{2}{c|}{F1} & \multicolumn{2}{c|}{Top-1} & \multicolumn{2}{c}{String-expert weight} \\
Benchmark & With & Without & With & Without & With & Without \\
\midrule
ParaNames, six languages & 75.9 & 75.2 & 73.6 & 72.7 & 25.8 & 6.4 \\
GeoNames, name only & 70.1 & 67.3 & 67.6 & 64.1 & 36.6 & 7.9 \\
GeoNames, name with country and region & 95.1 & 94.4 & 94.8 & 94.0 & 31.5 & 16.2 \\
Wikipedia articles to English & 99.5 & 99.5 & 99.5 & 99.5 & 9.9 & 14.9 \\
\bottomrule
\end{tabular}
\end{adjustbox}
\caption{EnsembleLink with and without transliteration, held-out test data (\%). ``String-expert weight'' is the share of the agreement weight on the sparse and Jaro-Winkler experts.}
\label{tab:translit}
\end{table}

Removing this transliteration step on the cross-script benchmarks costs little overall accuracy, at most about three points of F1 (Table~\ref{tab:translit}). On the cross-script name benchmarks, the string similarity experts without transliteration rarely agree with the consensus, and the ensemble shifts agreement weight to the embedding and reranker experts. On the Wikipedia articles, the string similarity experts receive little weight regardless of whether transliteration is first applied, and performance is unchanged. 

\subsection{Error Analysis}
\label{app:errors}

On the North Carolina voters task, almost every error occurs because the ensemble ranks a wrong candidate above a true record that is in the retrieved short list (Table~\ref{tab:linkage_errors}). On the names across languages task, retrieval misses a larger share of true records. On GeoNames, nearly all of the misses come from the name-only task, where a query and the true record share neither characters nor context. 

\begin{table}[!t]
\centering
\small
\adjustbox{max width=\textwidth}{\input{tables/tab_linkage_errors.tex}}
\caption{EnsembleLink errors on held-out test queries that have a counterpart. ``Not retrieved'' counts true records absent from the candidate pool, ``Ranked below a wrong record'' counts queries where another candidate scores higher, and ``Tied'' counts queries where the true record shares the highest score with another candidate. Top-1 accuracy counts tied queries as correct.}
\label{tab:linkage_errors}
\end{table}

\subsection{Match Confidence and Calibration}
\label{app:calibration}

EnsembleLink returns a similarity score rather than a probability. To provide a match confidence, which allows flagging uncertain matches or carrying linkage uncertainty into later downstream analyses, I average the rank of the margin between the top two candidates and the rank of the rerankers' own probability for the top candidate.

\begin{table}[!t]
\centering
\small
\adjustbox{max width=\textwidth}{\input{tables/tab_calibration.tex}}
\caption{Match confidence on held-out test queries that receive a proposed match. Discrimination is the area under the ROC curve for separating correct from incorrect matches. ECE is the expected calibration error of the confidence as returned (``Raw'') and after isotonic recalibration (``Recalibrated''). EnsembleLink returns a rank score, which has no raw ECE. Bold indicates the highest AUC and the lowest ECE in each panel.}
\label{tab:calibration}
\end{table}

On discrimination (Table~\ref{tab:calibration}), the EnsembleLink confidence comes within a hundredth of the best baseline on the North Carolina voters and leads by a wide distance across scripts. On calibration, fuzzylink probabilities understate fuzzylink accuracy across scripts, while the Fellegi-Sunter probabilities are well calibrated on the voters. Isotonic regression applied to each method leaves every match unchanged, because it only rescales the confidence; it also requires a small labeled sample. 

\subsection{One-to-One Matching}
\label{app:onetoone}

By default, EnsembleLink links each query to the highest-scoring candidate separately by query, which allows two queries to link to the same record. For one-to-one tasks, such as deduplication or linking the same people across two files, the scores can instead be used for single assignment. However, where many queries correctly share one record, such as variant spellings of the same place, a forced one-to-one assignment would link all but one of them to wrong records.

I test this as follows. I pool the candidate pairs for every query and sort them by the native EnsembleLink score. I first link the highest-scoring pair, setting aside that query and its matched record, and then continue down the sorted list, skipping any pair where either the query or the record has already been linked. This links each record to at most a single query. Queries whose best candidate has already been matched fall back to the next available, or are unmatched if none remain. On the North Carolina voters task, where each record has at most a single counterpart, this procedure leaves top-1 accuracy unchanged (98.04\% with each query linked independently; 98.00\% with one-to-one assignment).

\subsection{Computational Cost}
\label{app:scaling}

I calculate the computational cost of running the EnsembleLink package on a single NVIDIA A100 GPU (80 GB) on a record linkage task of North Carolina voters. I use name and birth year to join 1,000 voters to target data of between ten thousand and one million voter records, and between 100 and 3,000 voters to target data of 100,000 records. All target data contain the true matches (Table~\ref{tab:scaling}). I run each configuration in a new process, so that the total time includes loading the models. Embedding and indexing the target data takes the most time at every size.

\begin{table}[!t]
\centering
\small
\adjustbox{max width=\textwidth}{\input{tables/tab_scaling.tex}}
\caption{Total run time in minutes on one NVIDIA A100 GPU, including loading the models.}
\label{tab:scaling}
\end{table}

Because the package compares each query with every record, retrieval time grows with the number of queries times the size of the target data. Reranking time does not grow with the size of the target data, because each query contributes at most sixty candidate pairs, and so grows only with the number of queries. Target data need to be embedded only once, so linking a later batch of queries to the same target data adds only the retrieval and reranking time of the new queries.

\section{Ethnicity Inference}
\label{app:ethnicity}

\subsection{Prompt Templates}
\label{app:prompts}

The main text reports results for the default prompt defined below. For Gemini~3.0 Flash, which I use as the default model across tests, I set the reasoning effort to high, except in the Lebanese validation, where it runs with minimal reasoning. In the model panel, I instead run each model at the lowest reasoning setting. For every model whose reasoning can be explicitly set, I compare the model at its lowest reasoning setting and with reasoning set to high or enabled. For the remaining models. I compare the lowest setting with the provider default (Table~\ref{tab:reasoning}). High reasoning raises accuracy by about two points for Gemini~3.0 Flash and GPT-5.5, changes it by less than half a point for Claude Sonnet 4.6 and Qwen3.5-Plus, and lowers it by 12 to 40 points for GPT-5.4, GPT-5.4 mini, GPT-5.6 Luna, and GPT-6 Luna. As noted in the main text, these GPT model failures largely arise from over-prediction of the majority group when reasoning is enabled.

\begin{table}[!t]
\centering
\small
\begin{adjustbox}{max width=\textwidth}
\begin{tabular}{lllcccc}
\toprule
Model & Lowest setting & Second setting & Lowest & Second & $\Delta$ & $>$+names \\
\midrule
\multicolumn{7}{l}{\textit{Lowest vs.\ high or enabled reasoning}} \\
Gemini 3.0 Flash & Minimal & High & 77.1 & 78.9 & $+$1.8 & \checkmark~(4) \\
Gemini 3.8 Flash & Low & High & 78.1 & 78.6 & $+$0.5 & \checkmark~(4) \\
GPT-5.5 & None & High & 76.2 & 78.0 & $+$1.8 & \checkmark~(3) \\
DeepSeek-V4.1-Flash & Off & High (explicit) & 75.0 & 77.9 & $+$2.9 & \checkmark~(3) \\
DeepSeek-V4-Pro (0813) & Off & High (explicit) & 75.7 & 77.0 & $+$1.3 & \checkmark~(3) \\
Gemini 3.1 Flash Lite & Minimal & High & 73.7 & 76.9 & $+$3.2 & \checkmark~(3) \\
GPT-5.4 & None & High & 76.3 & 64.2 & $-$12.1 &  \\
GPT-6 Luna & None & High & 74.4 & 61.4 & $-$13.0 &  \\
GPT-5.6 Luna & None & High & 73.3 & 60.0 & $-$13.3 &  \\
GPT-5.4 mini & None & High & 73.2 & 33.0 & $-$40.2 &  \\
Claude Sonnet 4.6 & Default & Thinking on & 76.8 & 77.2 & $+$0.4 & \checkmark~(3) \\
Qwen3.5-Plus & Off & Reasoning on & 74.8 & 75.0 & $+$0.2 & \checkmark~(1) \\
\midrule
\multicolumn{7}{l}{\textit{Lowest vs.\ provider default (reasoning not set)}} \\
Gemini 3.5 Flash & Minimal & Provider default & 76.3 & 78.3 & $+$2.0 & \checkmark~(4) \\
Muse Spark 1.3 & Minimal & Provider default & 76.6 & 75.8 & $-$0.8 & \checkmark~(2) \\
GLM-5.3 & Minimal & Provider default & 74.8 & 75.4 & $+$0.6 & \checkmark~(1) \\
GLM-5-Turbo & Off & Provider default & 72.6 & 74.6 & $+$2.0 & \checkmark~(1) \\
DeepSeek-V4 & Off & Provider default & 75.7 & 74.5 & $-$1.2 &  \\
DeepSeek-V3.2 & Off & Provider default & 73.4 & 74.2 & $+$0.8 & \checkmark~(1) \\
Qwen3.8-2.4T & Minimal & Provider default & 63.4 & 65.6 & $+$2.2 &  \\
Muse Glimmer 30B & Minimal & Provider default & 57.0 & 62.2 & $+$5.2 &  \\
Muse Spark 1.2 & Minimal & Provider default & 42.5 & 48.3 & $+$5.8 &  \\
\bottomrule
\end{tabular}
\end{adjustbox}
\caption{Pooled accuracy (\%) over 2{,}000 voters in each of four states, at the lowest reasoning setting and at high, enabled, or default reasoning. Responses outside the four categories count as incorrect. $\Delta$ is the second setting minus the lowest, in percentage points, computed from the rounded values. A check mark ($>$+names) indicates that the model outperforms name-supplemented BISG in at least one state at the second setting, with the number of states in parentheses. Claude Opus 4.8, which has no reasoning setting, is not shown.}
\label{tab:reasoning}
\end{table}

The prompt templates that I test are as follows. Each was written in conjunction with Gemini~3.0 Flash, under the assumption that an LLM is more capable of writing a classification prompt for itself and other LLMs.

US Classification (Full Name + County):
\begin{verbatim}
You are an expert in American onomastics. Classify the race/ethnicity
of this person based on their full name and county of residence.

Full Name: {name}
County: {county}, {state}

Return ONLY one of these categories: WHITE, BLACK, HISPANIC, ASIAN, OTHER
Do not explain. Just return the single category.
\end{verbatim}

US Classification (Surname Only):
\begin{verbatim}
You are an expert in American onomastics. Classify the most likely
race/ethnicity associated with this surname.

Surname: {surname}

Return ONLY one of these categories: WHITE, BLACK, HISPANIC, ASIAN, OTHER
Do not explain. Just return the single category.
\end{verbatim}

Lebanon Classification (Gemini):
\begin{verbatim}
You are an expert in Lebanese onomastics. Classify the religious sect
of this Lebanese voter based on their name and region.

Name: {name}
Region: {region}

Return ONLY one of: SUNNI, SHIA, MARONITE, DRUZE, GREEK_ORTHODOX,
GREEK_CATHOLIC, ARMENIAN_ORTHODOX, OTHER

Do not explain your reasoning.
\end{verbatim}

Lebanon Classification (other models):
\begin{verbatim}
You are an expert in Lebanese onomastics. Classify the religious sect
of this Lebanese voter.

Name: {name}
Region: {region}

Return ONLY one of: SUNNI, SHIA, MARONITE, DRUZE, GREEK_ORTHODOX,
GREEK_CATHOLIC, ARMENIAN_ORTHODOX, OTHER
\end{verbatim}

India Classification:
\begin{verbatim}
You are an expert in Indian onomastics. Based on this person's name
and state, classify their most likely religion.

Name: {name}
State: {state}

Return ONLY one of: HINDU, MUSLIM, CHRISTIAN, SIKH, BUDDHIST, JAIN, OTHER
Do not explain. Just return the single category.
\end{verbatim}

India Caste Classification (four-way, Rajasthan sarpanch and Bihar land records):
\begin{verbatim}
Classify the caste category of this Indian person.
Name: {name}. District: {district}.
Return ONLY: GEN, OBC, SC, ST
\end{verbatim}

India Reserved-Constituency Classification (Scheduled Caste vs Scheduled Tribe):
\begin{verbatim}
Classify if this Indian Lok Sabha MP is Scheduled Caste (SC) or
Scheduled Tribe (ST). Name: {name}. State: {state}.
Return ONLY: SC or ST
\end{verbatim}

Aggregate Validation (one prompt per country, each with that country's own census
category list; two representative examples):
\begin{verbatim}
[Armenia]
Classify the ethnicity. Name: {name}. Region: {region}.
Return ONLY: ARMENIAN, RUSSIAN, YEZIDI, KURD, ASSYRIAN, GREEK, OTHER

[Chile]
Classify the ethnicity. Name: {name}. Region: {region}.
Return ONLY: MESTIZO, MAPUCHE, AYMARA, EUROPEAN, OTHER_INDIGENOUS, OTHER
\end{verbatim}

To test the sensitivity of results to the wording of the prompt, I compare six prompts with five models on the Florida panel sample of 2{,}000 voters (500 in each racial category). Besides the default prompt, these are: a minimal prompt with only the name and location, a detailed prompt with instructions on naming conventions, the default prompt without the ``expert in American onomastics'' framing, and the default prompt with five or with ten labeled examples (Table~\ref{tab:prompt_wording}).

\begin{table}[!t]
\centering
\small
\adjustbox{max width=\textwidth}{\input{tables/tab_prompt_wording.tex}}
\caption{Accuracy (\%) and responses outside the five categories by prompt, Florida panel sample (500 voters per racial category). Responses outside the categories count as incorrect. Model abbreviations as in Table~\ref{tab:ethnicity_accuracy}. Bold indicates the most accurate prompt for each model.}
\label{tab:prompt_wording}
\end{table}

Gemini~3.0 Flash varies in accuracy by about 3.6 points across the six prompts. Switching from the default prompt to the detailed prompt changes the accuracy of each model by less than two points. Neither prompt produces a model response outside of the five explicitly defined categories. Under the minimal prompt, in contrast, several models decline to classify some voters, or respond with a group classification outside of the five categories. This is highly model specific; for instance, GPT-4o does so for 1{,}842 of the 2{,}000 voters. The prompts with examples also lower the accuracy of GPT-4o, mostly through these invalid responses. It is therefore recommended that researchers explicitly state the task and the pre-defined acceptable categories in the prompt.

\subsection{Further Validation Results}
\label{app:validation}

\subsubsection{U.S. Officials}
\label{app:officials}

\citet{lee2025measuring} provide a data set of U.S.\ local elected officials with verified race labels. On 1,000 records drawn at random from their test set, requests to Gemini~3.0 Flash with the official's name and city achieve an accuracy of 90.4\% (Table~\ref{tab:leevelez}), exceeding the accuracies of BISG, fBISG, and the Lee and Velez combination of facial photographs and names. GPT-4o and GLM-5-Turbo both come within half a point of the photograph and name approach, and every model is at least as accurate as BISG. Unlike in the voter files, removing the city name lowers accuracy considerably for Gemini and GPT-4o for the public officials. Accuracy is higher here than in the Florida and North Carolina stratified voter file samples, because about 65\% of the officials are White and White names are easier to classify.

\begin{table}[!t]
\centering
\small
\begin{adjustbox}{max width=\textwidth}
\begin{tabular}{lccccc}
\toprule
Method & Gem. & 4o & DS4 & GLM5 & Q3.5 \\
\midrule
LLM (full name + city) & \textbf{90.4} & 88.3 & 84.7 & 88.5 & 86.0 \\
LLM (full name only) & 81.4 & 83.0 & 84.4 & \textbf{88.6} & 84.1 \\
\bottomrule
\end{tabular}
\end{adjustbox}
\caption[Lee and Velez validation]{Accuracy (\%) on 1,000 officials from the \citet{lee2025measuring} test set. Gem.\ and 4o are scored on one random sample, and DS4, GLM5, and Q3.5 on a second. Model abbreviations as in Table~\ref{tab:ethnicity_accuracy}. Lee and Velez report 88.8 for their combination of photographs and names, 85.4 for fBISG, 84.5 for BISG, and 81.9 for surname alone.}
\label{tab:leevelez}
\end{table}

In contrast to the voter files, memorization and recall of officials' races from the training data are a concern here. Local elected officials are public figures whose names, photos, and often race appear on campaign sites, local news platforms, Wikipedia, and information aggregators like Ballotpedia. It is possible then that an LLM could get these inferences correct by recalling the person rather than inferring their race based on their name. However, Gemini's margin over BISG is smaller on the officials than on the voters, whereas simple recall would tend to inflate that margin. This general problem for benchmarking is redressed with the synthetic voter name result in Appendix~\ref{app:synthetic}.

\subsubsection{Florida and North Carolina}
\label{app:flnc}

The differences in accuracy between the LLM and BISG in Florida and North Carolina are statistically significant under a paired McNemar test, which compares the two methods' successes and failures at the record level. Against standard BISG, the test rejects equal accuracy at $p<10^{-200}$ in both states ($\chi^2_1 = 1141$ in Florida and 985 in North Carolina), and against name-supplemented BISG at $p<10^{-40}$ ($\chi^2_1 = 275$ and 194). 

BISG is less accurate in these samples than in other published validations, because the samples used here are stratified samples that weight each racial category equally, whereas about 60\% of voters in both states are White. BISG classifies White voters much better than minority voters, so weighting the groups equally lowers BISG accuracy. For reference, fBISG scores 68.0\% in Florida and 66.6\% in North Carolina with surname and geography, and 77.6\% in both states with first and middle names added. I report recall and F1 by racial category for each model and for BISG in Tables~\ref{tab:ethnicity_recall} and~\ref{tab:ethnicity_prf}.

\begin{table}[!t]
\centering
\resizebox{\textwidth}{!}{
\begin{tabular}{l|cccccc|cc|cccccc|cc}
\toprule
 & \multicolumn{8}{c|}{Florida} & \multicolumn{8}{c}{North Carolina} \\
Race & Gem. & 4o & 4.1-m & DS4 & GLM5 & Q3.5 & BISG & +nm & Gem. & 4o & 4.1-m & DS4 & GLM5 & Q3.5 & BISG & +nm \\
\midrule
White & 88.4 & 88.5 & \textbf{90.6} & 86.8 & 90.4 & 86.8 & 86.8 & 91.5 & 84.4 & 88.2 & \textbf{90.5} & 82.4 & 89.2 & 87.3 & 85.0 & 92.9 \\
Black & \textbf{81.7} & 72.2 & 48.9 & 62.8 & 63.0 & 63.0 & 48.8 & 67.0 & \textbf{80.0} & 68.3 & 52.1 & 71.4 & 63.3 & 64.1 & 53.1 & 65.2 \\
Hispanic & 92.3 & 91.6 & 91.4 & 92.2 & 90.0 & \textbf{92.6} & 88.8 & 93.1 & 84.9 & 85.7 & 83.8 & 84.7 & 81.5 & \textbf{85.8} & 77.0 & 81.2 \\
Asian & \textbf{72.8} & 61.6 & 51.2 & 64.4 & 52.7 & 57.2 & 48.5 & 60.4 & \textbf{84.5} & 79.6 & 70.2 & 81.6 & 73.4 & 76.2 & 60.5 & 74.4 \\
\hdashline
Average & \textbf{83.8} & 78.5 & 70.5 & 76.6 & 74.0 & 74.9 & 68.2 & 78.0 & \textbf{83.5} & 80.5 & 74.1 & 80.0 & 76.8 & 78.3 & 68.9 & 78.4 \\
\bottomrule
\end{tabular}}
\caption[Recall by racial category]{Recall by racial category (\%), full name and geography. +nm is BISG with first and middle name supplements \citep{imai2022addressing}. Bold indicates the most accurate LLM in each row.}
\label{tab:ethnicity_recall}
\end{table}

\begin{table}[!t]
\centering
\resizebox{\textwidth}{!}{
\begin{tabular}{l|ccccccc|ccccccc}
\toprule
 & \multicolumn{7}{c|}{Florida} & \multicolumn{7}{c}{North Carolina} \\
 & Gem. & 4o & 4.1-m & DS4 & GLM5 & Q3.5 & BISG & Gem. & 4o & 4.1-m & DS4 & GLM5 & Q3.5 & BISG \\
\midrule
\multicolumn{15}{l}{\textit{F1 by racial category}} \\
White & 79.8 & 76.6 & 71.6 & 74.5 & 73.6 & 74.2 & 65.2 & 76.4 & 74.7 & 70.1 & 73.0 & 72.1 & 73.0 & 62.2 \\
Black & 83.9 & 78.7 & 63.1 & 72.3 & 72.8 & 72.2 & 60.3 & 80.0 & 74.7 & 63.8 & 73.6 & 71.3 & 71.4 & 58.8 \\
Hispanic & 88.3 & 86.4 & 83.7 & 82.4 & 85.5 & 82.8 & 80.2 & 89.0 & 88.4 & 87.0 & 87.2 & 86.6 & 87.4 & 84.4 \\
Asian & 83.8 & 75.7 & 67.4 & 77.3 & 68.7 & 72.5 & 64.9 & 91.1 & 88.2 & 82.2 & 88.8 & 84.4 & 86.1 & 74.8 \\
\hdashline
Macro-F1 & \textbf{83.9} & 79.3 & 71.4 & 76.6 & 75.2 & 75.4 & 67.7 & \textbf{84.1} & 81.5 & 75.8 & 80.7 & 78.6 & 79.5 & 70.1 \\
Macro-precision & \textbf{85.5} & 83.5 & 80.9 & 80.4 & 82.2 & 80.8 & 75.6 & \textbf{85.6} & 84.4 & 82.3 & 82.2 & 83.5 & 82.8 & 76.6 \\
\bottomrule
\end{tabular}}
\caption{Per-class F1 and macro-averaged precision and F1 (\%), full name and geography. BISG uses surname and county.}
\label{tab:ethnicity_prf}
\end{table}

\subsubsection{Lebanon}
\label{app:lebanon}

In Lebanon, accuracy is higher for sects that have more distinctive names (Armenian Orthodox and Shia) and lower for the two Christian sects whose naming conventions have converged on one another (Greek Orthodox and Greek Catholic; Table~\ref{tab:lebanon}). 

\begin{table}[!t]
\centering
\small
\begin{adjustbox}{max width=\textwidth}
\begin{tabular}{lcccccc}
\toprule
Sect & Gem. & 4o & 4.1-m & DS4 & GLM5 & Q3.5 \\
\midrule
Armenian Orthodox & \textbf{97.2} & 95.6 & 72.4 & 91.4 & 79.0 & 95.8 \\
Shia & 85.8 & \textbf{90.8} & 81.2 & 89.8 & 84.4 & 64.8 \\
Maronite & 72.6 & 63.6 & 43.2 & \textbf{75.0} & 72.6 & 54.6 \\
Sunni & 77.4 & 51.4 & \textbf{82.0} & 46.0 & 51.0 & 76.6 \\
Druze & \textbf{55.4} & 49.6 & 30.4 & 25.8 & 0.6 & 28.2 \\
Greek Orthodox & \textbf{43.2} & 31.4 & 4.2 & 9.0 & 16.6 & 34.2 \\
Greek Catholic (Melkite) & \textbf{34.2} & 6.0 & 1.4 & 2.4 & 8.2 & 9.8 \\
\hdashline
Overall & \textbf{66.5} & 55.5 & 45.0 & 48.5 & 44.6 & 52.0 \\
\bottomrule
\end{tabular}
\end{adjustbox}
\caption{Accuracy by religious sect (\%), Lebanon. Model abbreviations as in Table~\ref{tab:ethnicity_accuracy}. Gemini runs with minimal reasoning. The GLM-5-Turbo column reflects the hosted model as of October 2026.}
\label{tab:lebanon}
\end{table}

I cannot run BISG truly out-of-sample in Lebanon, because there are no sect-surname tables that exist outside of the voter file. I therefore derive these tables from the voter file itself and predict the sect of each voter with that voter left out of the estimation. With the same regions whose names are provided to the LLM alongside the voters' names, this implementation of BISG reaches about 63\% accuracy, which is close to Gemini~3.0 Flash and better than the other LLMs. BISG is even more accurate with the more granular district geography. BISG is strongest on the geographically concentrated sects (Druze and Greek Catholic) and the LLMs on the sects with more distinctive names (Armenian Orthodox and Shia). I interpret these results to suggest that the best LLMs perform marginally better than BISG with the same regional information in a setting where the requisite data for BISG is not ``truly'' known. 

\subsubsection{India}
\label{app:india}

For India, I report the binary SC versus ST classification of Lok Sabha MPs from reserved constituencies in Table~\ref{tab:india_scst}.

\begin{table}[!t]
\centering
\small
\begin{adjustbox}{max width=\textwidth}
\begin{tabular}{lccccc}
\toprule
Category & Gem. & 4o & DS4 & GLM5 & Q3.5 \\
\midrule
Scheduled Caste (SC) & \textbf{100.0} & 98.8 & 89.3 & 70.2 & 76.2 \\
Scheduled Tribe (ST) & \textbf{97.9} & 87.2 & 72.3 & \textbf{97.9} & 93.6 \\
\midrule
Overall & \textbf{99.2} & 94.7 & 83.2 & 80.2 & 82.4 \\
\bottomrule
\end{tabular}
\end{adjustbox}
\caption{Accuracy (\%) classifying the 131 Lok Sabha MPs elected from reserved constituencies in 2024 as Scheduled Caste or Scheduled Tribe. Model abbreviations as in Table~\ref{tab:ethnicity_accuracy}.}
\label{tab:india_scst}
\end{table}

In the four-way caste classification (Table~\ref{tab:india_caste}),\footnote{Data provided by Gaurav Sood.} Gemini~3.0 Flash achieves an accuracy of 75.3\% on the Bihar land records but only 57.7\% on the Rajasthan village heads. Many Bihar names include a caste surname, such as the Scheduled Caste surnames Paswan, Chamar, and Dusadh (93.3\% accuracy for Scheduled Castes), whereas Rajasthan names appear to carry fewer consistent markers of caste.

\begin{table}[!t]
\centering
\small
\resizebox{\textwidth}{!}{%
\begin{tabular}{lcc|cc|cc|cc}
\toprule
 & \multicolumn{2}{c|}{Gemini~3.0 Flash} & \multicolumn{2}{c|}{DeepSeek-V4} & \multicolumn{2}{c|}{GLM-5-Turbo} & \multicolumn{2}{c}{Qwen3.5-Plus} \\
Category & Rajasthan & Bihar & Rajasthan & Bihar & Rajasthan & Bihar & Rajasthan & Bihar \\
\midrule
Scheduled Caste (SC) & \textbf{56.0} & \textbf{93.3} & 33.3 & 72.0 & 2.7 & 56.0 & 5.3 & 49.3 \\
Scheduled Tribe (ST) & 66.7 & 68.0 & 77.3 & 85.3 & 50.7 & 50.7 & \textbf{97.3} & \textbf{97.3} \\
Other Backward Class (OBC) & \textbf{61.3} & \textbf{68.0} & 24.0 & 60.0 & 18.7 & 45.3 & 4.0 & 22.7 \\
General (unreserved) & 46.7 & 72.0 & 21.3 & 38.7 & \textbf{86.7} & \textbf{82.7} & 2.7 & 37.3 \\
\midrule
Overall & \textbf{57.7} & \textbf{75.3} & 39.0 & 64.0 & 39.7 & 58.7 & 27.3 & 51.7 \\
\bottomrule
\end{tabular}}
\caption{Accuracy for four-way caste classification (\%). Rajasthan: elected village heads (\textit{sarpanch}) with official category labels. Bihar: land records with recorded \textit{jati}.}
\label{tab:india_caste}
\end{table}

The other models are less accurate than Gemini except on Scheduled Tribes (DeepSeek-V4 and Qwen3.5-Plus) and the general category (GLM-5-Turbo). These results are consistent with the information saturation discussed in the main paper: perhaps mechanically, LLMs are only able to infer ethnic group membership from names in settings where names are representative of the respective ethnic groups.

\subsection{Model Panel: Identifiers, Providers, and Licenses}
\label{app:panel_models}

I list the exact API identifier, reasoning setting, and license of each model used in the panel (Table~\ref{tab:durability}) in Table~\ref{tab:panel_models}. I ran twenty of the models in June 2026 and fifteen in September and October 2026. I report the time of each API call in the replication materials. I called most models through OpenRouter for uniformity in the structured output. The exceptions are the GPT-5 and GPT-6 models, which were called through the OpenAI Responses API, and Gemini 3.8 Flash, which I called through the Gemini API. 

I use the lowest reasoning setting possible for every model, either disabling reasoning entirely where possible or otherwise setting it to the lowest reasoning level. GLM-5.3, the Muse models, and Qwen3.8-2.4T do not allow for reasoning to be fully disabled, while Gemini 3.8 Flash does not accept a reasoning level below ``low.'' These models thus still reason before outputting the inferred ethnic classification, though only briefly. 

I ran every panel model through a hosted service and did not rerun any model locally. Open-weight models can nonetheless be downloaded and run on personal devices. I group the hosted commercial models of model families that are generally open-weight, but whose weights are not released (GLM-5-Turbo, Qwen3.5-Plus, Qwen3.5-Flash, Qwen3.7-Plus, and Qwen3.8-Flash), together with the proprietary models.

I count responses other than the specified categories as incorrect. These responses, at most 27 of 2{,}000 records for any given model and state, are mostly empty. Because the panel uses a fixed stratified sample of 2{,}000 records per state instead of the 10{,}000-record validation samples, panel accuracy differs slightly from the main validation results. The main results also use Gemini~3.0 Flash with high reasoning (82.0\% in Florida in the panel at that setting and 83.8\% in the main validation).

\begin{table}[!t]
\centering
\scriptsize
\setlength{\tabcolsep}{4pt}
\begin{adjustbox}{max width=\textwidth}
\begin{tabular}{llll}
\toprule
Model & API identifier & Reasoning & License \\
\midrule
\multicolumn{4}{l}{\textit{Proprietary (API-only)}} \\
Gemini~3.0 Flash & \texttt{google/gemini-3-flash-preview} & minimal & Proprietary \\
Gemini 3.5 Flash & \texttt{google/gemini-3.5-flash} & minimal & Proprietary \\
Gemini 3.8 Flash & \texttt{gemini-3.8-flash} (Gemini API) & low & Proprietary \\
GPT-5.5 & \texttt{gpt-5.5} (OpenAI Responses) & none & Proprietary \\
Claude Opus 4.8 & \texttt{anthropic/claude-opus-4.8} & default & Proprietary \\
Claude Sonnet 4.6 & \texttt{anthropic/claude-sonnet-4.6} & default & Proprietary \\
Gemini 3.1 Flash Lite & \texttt{google/gemini-3.1-flash-lite-preview} & minimal & Proprietary \\
Muse Spark 1.3 & \texttt{meta/muse-spark-1.3} & minimal & Proprietary \\
GPT-4o & \texttt{openai/gpt-4o} & default & Proprietary \\
Qwen3.7-Plus & \texttt{qwen/qwen3.7-plus} & off & Proprietary \\
GLM-5-Turbo & \texttt{z-ai/glm-5-turbo} & off & Proprietary \\
Qwen3.5-Plus & \texttt{qwen/qwen3.5-plus-02-15} & off & Proprietary \\
GPT-4.1 mini & \texttt{openai/gpt-4.1-mini} & default & Proprietary \\
Qwen3.8-Flash & \texttt{qwen/qwen3.8-flash} & off & Proprietary \\
Qwen3.5-Flash & \texttt{qwen/qwen3.5-flash-02-23} & off & Proprietary \\
GPT-5.4 & \texttt{gpt-5.4} (OpenAI Responses) & none & Proprietary \\
GPT-6 Luna & \texttt{gpt-6-luna} (OpenAI Responses) & none & Proprietary \\
GPT-5.6 Luna & \texttt{gpt-5.6-luna} (OpenAI Responses) & none & Proprietary \\
Muse Spark 1.2 & \texttt{meta/muse-spark-1.2} & minimal & Proprietary \\
GPT-5.4 mini & \texttt{gpt-5.4-mini} (OpenAI Responses) & none & Proprietary \\
\hdashline
\multicolumn{4}{l}{\textit{Open-weight (downloadable, version-pinnable)}} \\
DeepSeek-V4.1-Flash & \texttt{deepseek/deepseek-v4.1-flash} & off & MIT \\
DeepSeek-V4-Pro (0813) & \texttt{deepseek/deepseek-v4-pro-0813} & off & MIT \\
GLM-5.3 & \texttt{z-ai/glm-5.3} & minimal & GLM-5.3 license (custom) \\
Qwen3.5-397B & \texttt{qwen/qwen3.5-397b-a17b} & off & Apache-2.0 \\
DeepSeek-V4 & \texttt{deepseek/deepseek-v4-pro} & off & MIT \\
DeepSeek-V3.2 & \texttt{deepseek/deepseek-v3.2} & off & MIT \\
Llama-4 Maverick & \texttt{meta-llama/llama-4-maverick} & default & Llama 4 Community \\
Qwen3.5-27B & \texttt{qwen/qwen3.5-27b} & off & Apache-2.0 \\
Qwen3.8-27B & \texttt{qwen/qwen3.8-27b} & off & Apache-2.0 \\
Qwen3.6-35B & \texttt{qwen/qwen3.6-35b-a3b} & off & Apache-2.0 \\
Qwen3.8-2.4T & \texttt{qwen/qwen3.8-2.4t-a95b} & minimal & Qwen3.8-Max license (custom) \\
Qwen3.5-122B & \texttt{qwen/qwen3.5-122b-a10b} & off & Apache-2.0 \\
Qwen3.5-35B & \texttt{qwen/qwen3.5-35b-a3b} & off & Apache-2.0 \\
Muse Glimmer 30B & \texttt{meta/muse-glimmer-30b} & minimal & Apache-2.0 \\
Qwen3.5-9B & \texttt{qwen/qwen3.5-9b} & off & Apache-2.0 \\
\bottomrule
\end{tabular}
\end{adjustbox}
\caption{API identifier, reasoning setting, and license for every model in Table~\ref{tab:durability}. Models without a noted API ran through OpenRouter. ``Off'' sets \texttt{reasoning.enabled=false}; ``none,'' ``minimal,'' and ``low'' are the lowest effort levels accepted by each API; and ``default'' indicates the provider default, with no reasoning requested.}
\label{tab:panel_models}
\end{table}

\subsection{Aggregate Validation}
\label{app:aggregate}

I scraped or downloaded the full voter files for Uganda, Nepal, Armenia, Chile, and Costa Rica from the websites of electoral commissions or from online archives, and for India from \citet{sood2018parsed}. I run individual-level inference using each model with respect to the listed groups and then aggregate the national population share of each group across individuals, then compare this proportion to the proportions reported in the corresponding census or official government reporting (Table~\ref{tab:aggregate_full}). 

In India, the aggregate Hindu and Muslim shares from Gemini~3.0 Flash are within 0.5 ppts of the census. Armenian names are distinctive by group, allowing every model to estimate the Armenian share almost exactly, and Gemini also recovers the Yezidi minority. Indigenous Costa Ricans are difficult for every model to detect, and their names indeed appear to overlap consistently with names of the majority group. In Uganda, the models estimate the smaller groups accurately, though they tend to overestimate the three largest ethnic groups. This result is expected if the models assign ambiguous names to the largest ethnic categories. 

\begin{table}[!t]
\centering
\small
\begin{adjustbox}{max width=\textwidth}
\begin{tabular}{llc|ccccc}
\toprule
Country & Category & Census & Gem. & 4o & DS4 & GLM5 & Q3.5 \\
\midrule
India & Hindu & 79.8 & \textbf{80.2} & 81.0 & 83.4 & 82.5 & 65.6 \\
India & Muslim & 14.2 & \textbf{13.9} & 13.1 & 11.2 & 12.5 & 12.1 \\
India & Christian & 2.3 & \textbf{2.1} & 0.8 & 1.3 & 0.8 & 1.4 \\
India & Sikh & 1.7 & \textbf{2.2} & 3.0 & 2.7 & 2.7 & 3.6 \\
India & Buddhist & 0.7 & \textbf{0.2} & 0.2 & 0.2 & 0.2 & 0.2 \\
India & Jain & 0.4 & 0.3 & \textbf{0.4} & 0.8 & 0.4 & 0.4 \\
India & Other & 0.9 & 1.0 & 1.4 & 0.4 & \textbf{0.9} & 16.7 \\
India & \textit{Avg. error} & -- & \textbf{0.3} & 0.9 & 1.4 & 1.1 & 5.1 \\
\hdashline
Uganda & Baganda & 16.5 & 24.7 & \textbf{19.8} & 26.7 & 31.9 & 22.3 \\
Uganda & Banyankole & 9.6 & 17.1 & \textbf{11.6} & 16.4 & 15.1 & 15.2 \\
Uganda & Basoga & 8.8 & 10.5 & \textbf{7.8} & 15.3 & 11.9 & 12.1 \\
Uganda & Iteso & 7.0 & 8.1 & 4.0 & \textbf{7.1} & 8.0 & 8.3 \\
Uganda & Langi & 6.1 & 7.1 & 8.6 & \textbf{6.7} & 2.2 & 3.8 \\
Uganda & Bagisu & 4.9 & 5.4 & 2.0 & \textbf{4.9} & 2.8 & 3.2 \\
Uganda & Acholi & 4.4 & 4.8 & \textbf{4.4} & 5.9 & 4.7 & 4.9 \\
Uganda & Lugbara & 4.2 & 3.4 & 3.2 & 3.2 & 3.4 & \textbf{4.5} \\
Uganda & Batooro & 3.2 & 3.6 & 2.8 & 1.2 & \textbf{3.0} & 3.8 \\
Uganda & Banyoro & 3.0 & \textbf{3.1} & 2.0 & 5.6 & 2.5 & 3.2 \\
Uganda & \textit{Avg. error} & -- & 2.2 & \textbf{1.7} & 3.1 & 3.3 & 2.2 \\
\hdashline
Nepal & Chhetri & 16.6 & 20.8 & 19.2 & 36.3 & 49.8 & \textbf{17.8} \\
Nepal & Brahmin & 12.2 & 16.2 & 15.8 & 16.3 & 7.6 & \textbf{12.9} \\
Nepal & Magar & 7.1 & 6.1 & 6.4 & 4.7 & \textbf{7.1} & 6.4 \\
Nepal & Tharu & 6.6 & \textbf{7.6} & 8.2 & 4.4 & 1.5 & 5.1 \\
Nepal & Tamang & 5.8 & \textbf{6.1} & 5.4 & 5.5 & 3.9 & 4.7 \\
Nepal & Newar & 5.0 & 5.7 & 6.2 & 4.0 & 1.0 & \textbf{5.3} \\
Nepal & Kami & 4.8 & 6.2 & \textbf{4.8} & 1.7 & 2.6 & 2.0 \\
Nepal & Yadav & 4.0 & 3.6 & 5.6 & 4.9 & \textbf{4.0} & 13.1 \\
Nepal & Rai & 2.3 & 3.2 & 3.4 & 3.5 & 3.5 & \textbf{3.1} \\
Nepal & Gurung & 2.0 & \textbf{2.3} & 2.8 & 2.4 & 2.4 & 2.4 \\
Nepal & Limbu & 1.5 & \textbf{1.6} & 1.2 & 1.1 & 1.1 & 1.3 \\
Nepal & \textit{Avg. error} & -- & 1.3 & \textbf{1.3} & 3.2 & 4.8 & 1.7 \\
\hdashline
Armenia & Armenian & 98.1 & \textbf{98.2} & 99.2 & 99.9 & 100.0 & 98.8 \\
Armenia & Yezidi & 1.2 & \textbf{1.0} & 0.0 & 0.0 & 0.0 & 0.2 \\
Armenia & Russian & 0.5 & 0.8 & 0.8 & 0.0 & 0.0 & \textbf{0.5} \\
Armenia & \textit{Avg. error} & -- & \textbf{0.2} & 0.9 & 1.2 & 1.2 & 0.6 \\
\hdashline
Chile & Mestizo & 87.2 & 90.0 & 82.2 & 93.0 & \textbf{87.7} & 91.3 \\
Chile & Mapuche & 9.9 & 4.0 & 5.2 & 1.7 & \textbf{9.2} & 3.0 \\
Chile & European & 1.6 & 2.6 & 5.0 & \textbf{1.7} & 1.2 & 4.0 \\
Chile & Aymara & 0.7 & 1.8 & 2.0 & 2.0 & 1.7 & \textbf{1.0} \\
Chile & \textit{Avg. error} & -- & 2.7 & 3.6 & 3.8 & \textbf{0.7} & 3.4 \\
\hdashline
Costa Rica & White or mestizo & 83.6 & 95.8 & \textbf{81.2} & 91.8 & 91.6 & 96.5 \\
Costa Rica & Afro-Carib. & 7.8 & 3.6 & 8.4 & 6.8 & \textbf{7.5} & 1.9 \\
Costa Rica & Indigenous & 2.4 & 0.2 & 0.0 & \textbf{0.4} & 0.0 & 0.0 \\
Costa Rica & Asian & 0.2 & \textbf{0.4} & 0.4 & 0.0 & 0.0 & 0.0 \\
Costa Rica & \textit{Avg. error} & -- & 4.7 & \textbf{1.4} & 2.9 & 2.7 & 5.4 \\
\bottomrule
\end{tabular}
\end{adjustbox}
\caption{Inferred population shares and census shares (\%). Uganda and Nepal show the largest groups only. Bold indicates the estimate closest to the census. Model abbreviations as in Table~\ref{tab:ethnicity_accuracy}.}
\label{tab:aggregate_full}
\end{table}

\subsection{Distillation into Small Models}
\label{app:distillation}

One key question for the use of LLMs in ethnicity inference is how to scale to large samples or full populations. Proprietary models can be costly to run on millions of names, and open-weight models can more often be inaccurate or costly to self-host. As an alternative for inference at scale, a researcher can label a small sample of names with a frontier LLM, fine-tune a small open-weight model on those labels, and then run the fine-tuned model locally at no marginal cost (Table~\ref{tab:distillation}). I showcase this using both small open-weight LLMs and a logistic classifier. 

\begin{table}[!t]
\centering
\small
\begin{adjustbox}{max width=\textwidth}
\begin{tabular}{llcccc}
\toprule
Model & Dataset & Teacher & Base & FT (LLM) & FT (true) \\
\midrule
Qwen3-4B & Florida & 83.3 & 72.0 & 78.7 & 78.3 \\
Qwen3-1.7B & Florida & 83.3 & 60.3 & 79.7 & 78.3 \\
Qwen3-0.6B & Florida & 83.3 & 7.3 & 75.7 & 76.0 \\
Logistic classifier & Florida & 83.3 & --- & 72.7 & 70.3 \\
\hdashline
Qwen3-4B & North Carolina & 71.7 & 63.7 & 68.0 & 70.3 \\
Qwen3-1.7B & North Carolina & 71.7 & 64.0 & 67.7 & 72.0 \\
Qwen3-0.6B & North Carolina & 71.7 & 18.0 & 68.0 & 66.0 \\
Logistic classifier & North Carolina & 71.7 & --- & 68.3 & 66.7 \\
\hdashline
Qwen3-4B & Lebanon & 76.0 & 14.0 & 62.0 & 61.0 \\
Qwen3-1.7B & Lebanon & 76.0 & 28.0 & 57.5 & 58.0 \\
Qwen3-0.6B & Lebanon & 76.0 & 4.5 & 53.5 & 48.5 \\
Logistic classifier & Lebanon & 76.0 & --- & 40.0 & 39.5 \\
\bottomrule
\end{tabular}
\end{adjustbox}
\caption{Accuracy (\%) of small models trained on LLM labels: three Qwen3 models fine-tuned with LoRA and a logistic classifier on multilingual-e5-large name embeddings. Base is accuracy before fine-tuning and Teacher the accuracy of Gemini~3.0 Flash. FT (LLM) trains on the Gemini labels and FT (true) on the true labels.}
\label{tab:distillation}
\end{table}

After fine-tuning on 1,200 names labeled by Gemini~3.0 Flash, costing only a few cents, even Qwen3-0.6B, which has almost no zero-shot ability, improves from 7.3\% to 75.7\% accuracy on the Florida voters task. In the settings that I test, training on the true labels instead of the labels from Gemini~3.0 Flash only changes accuracy by about two points on average, sometimes positively and sometimes negatively. Distillation from a small number of samples therefore enables inference on whole populations without true labels, usually at some loss of accuracy relative to the model that is used to generate those labels.  

\subsection{Discrimination}
\label{app:probrank}

Downstream analyses drawing on surname inferences rely on calibrated probabilities. Calibration concerns whether the probability of group membership corresponding to a group prediction matches the accuracy of the model (Section~\ref{subsec:uncertainty}). For instance, if a model assigns a probability of $p$\% to an individual's membership in a group, the prediction should be correct $p$\% of the time. A separate question from that of calibration is whether the probabilities rank the members of a group above everyone else, which allows the use of a threshold applied to the prediction scores for determining discrete group membership. For each racial category, I compute one-versus-rest precision and recall, scoring each person by the probability that a model assigns to the first token of the category and, for BISG, by the BISG posterior. Only some models return such probabilities; among proprietary models, only the OpenAI models return them. Among open-weight models, only DeepSeek-V3.2 and Llama-4 Maverick return them through OpenRouter. For the other open models, predicted probabilities come from the distilled classifier trained in Section~\ref{subsec:uncertainty}.

\begin{figure}[!t]
\centering
\includegraphics[width=\linewidth]{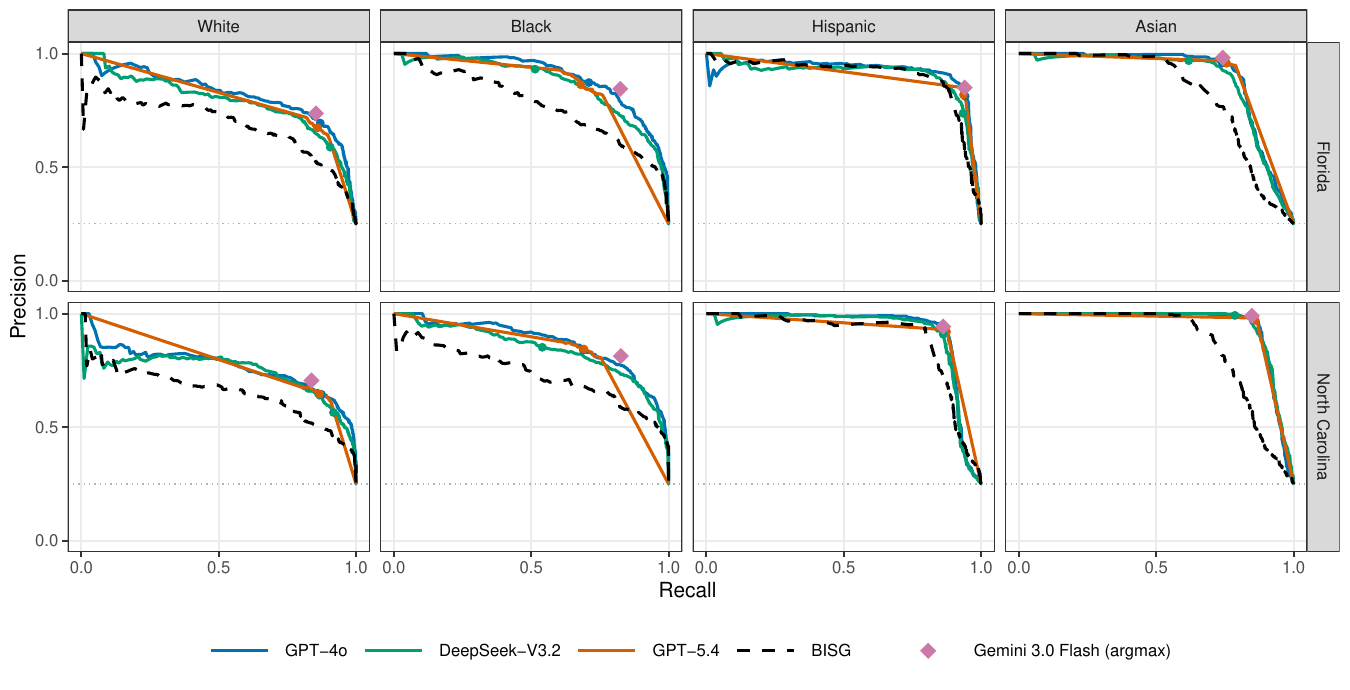}
\caption{One-versus-rest precision and recall by racial category, scoring each person by the first-token probability of the category and by the BISG posterior. The dotted line indicates the share of each category in the sample. Circles mark the most probable category of each model, and diamonds the precision and recall of Gemini~3.0 Flash, which returns no token probabilities.}
\label{fig:probrank}
\end{figure}

I plot precision-recall curves for a proprietary model (GPT-4o), a proprietary reasoning model (GPT-5.4), an open-weight model (DeepSeek-V3.2), and BISG (Figure~\ref{fig:probrank}), and report average precision for every model in the panel that returns probabilities (Table~\ref{tab:probrank}). GPT-4o, GPT-4.1 mini, DeepSeek-V3.2, and Llama-4 Maverick match or exceed the average precision of standard BISG in every category in both states.

\begin{table}[!t]
\centering
\small
\setlength{\tabcolsep}{5pt}
\begin{adjustbox}{max width=\textwidth}
\begin{tabular}{llcccccc}
\toprule
State & Category & GPT-4o & GPT-4.1-mini & DeepSeek-V3.2 & Llama-4 & GPT-5.4 & BISG \\
\midrule
Florida & White & \textbf{0.82} & 0.75 & 0.79 & 0.70 & 0.67 & 0.69 \\
 & Black & \textbf{0.90} & 0.81 & 0.87 & 0.81 & 0.76 & 0.77 \\
 & Hispanic & \textbf{0.92} & 0.91 & 0.91 & 0.90 & 0.82 & 0.90 \\
 & Asian & \textbf{0.90} & 0.85 & 0.88 & 0.84 & 0.82 & 0.83 \\
North Carolina & White & \textbf{0.77} & 0.69 & 0.75 & 0.65 & 0.63 & 0.65 \\
 & Black & \textbf{0.87} & 0.81 & 0.84 & 0.82 & 0.72 & 0.73 \\
 & Hispanic & \textbf{0.93} & 0.91 & 0.92 & 0.92 & 0.85 & 0.89 \\
 & Asian & \textbf{0.95} & 0.93 & \textbf{0.95} & 0.91 & 0.88 & 0.86 \\
\bottomrule
\end{tabular}
\end{adjustbox}
\caption{Average precision by state and racial category for the models that return probabilities and for BISG. Bold indicates the highest value in each row.}
\label{tab:probrank}
\end{table}

\begin{figure}[!t]
\centering
\includegraphics[width=.62\linewidth]{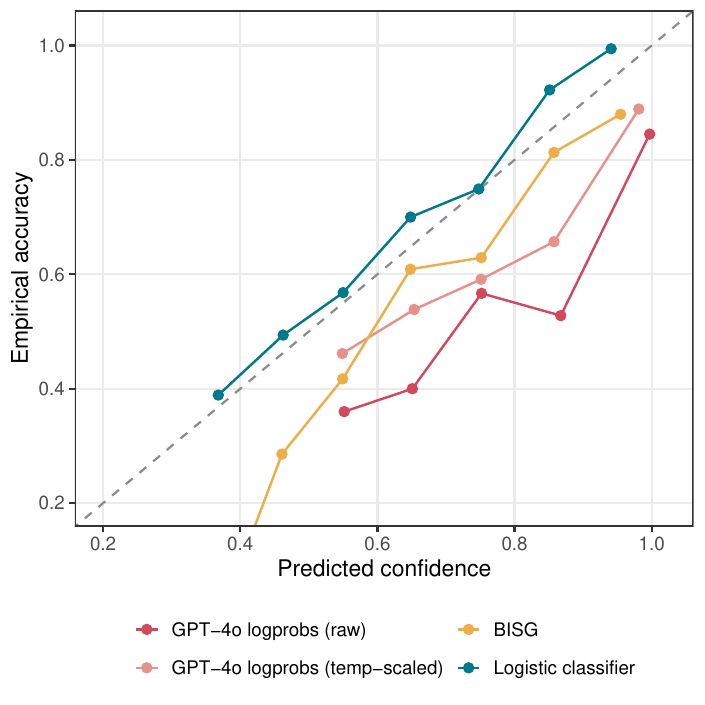}
\caption{Reliability diagram (predicted confidence against accuracy) for North Carolina voters, corresponding to the Florida figure in Section~\ref{subsec:uncertainty}.}
\label{fig:reliability_nc}
\end{figure}

The reliability diagram in Section~\ref{subsec:uncertainty}, which plots predicted confidence against empirical accuracy, draws on a sample of Florida voters. I repeat the same procedure for North Carolina in Figure~\ref{fig:reliability_nc}.

\subsection{Multiple-Imputation Intervals}
\label{app:mi_intervals}

Assigning each person to a single group before estimating group differences in an outcome, such as partisanship, ignores the uncertainty in the classification. Multiple imputation from the probabilities of a classifier allows us to measure this uncertainty, but only conditional on that classifier, and so does not by itself provide valid confidence intervals for the true racial gap in a given outcome. I draw 500 complete sets of race from the probabilities of the logistic classifier, estimate each racial gap in every draw, and combine the estimates using Rubin's rules. I do this for the same three outcomes and two states as in Appendix~\ref{app:birdie_llm}. Each comparison is the gap between a minority group and White voters for a given outcome in a single state. 

\begin{table}[!t]
\centering
\small
\begin{adjustbox}{max width=\textwidth}
\begin{tabular}{lcc}
\toprule
 & Geographic gaps & Party gap \\
\midrule
Conditional MI total std.\ error & 1.7 & 1.8 \\
Classifier plug-in $\lvert$bias$\rvert$ & 6.6 & 24.5 \\
Categorical LLM $\lvert$bias$\rvert$ & 2.1 & 10.8 \\
Interval contains true gap & 5/12 & 0/6 \\
\bottomrule
\end{tabular}
\end{adjustbox}
\caption{Multiple-imputation uncertainty in the gap between minority and White voters, conditional on the fitted logistic classifier. Bias is the absolute distance of each estimate from the gap measured with true race.}
\label{tab:mi_intervals}
\end{table}

I define ``geographic'' outcomes as those which vary strongly with geography, specifically frequent turnout (voting in at least half of the general elections between 2012 and 2018) and living in a high-income neighborhood (the voter's ZCTA has a median household income above the median). The intervals are narrow, with a standard error near 1.7 ppts on the geographic outcomes. However, the intervals contain the true racial gap in only five of the twelve geographic comparisons, and in none of the six comparisons of party registration (Table~\ref{tab:mi_intervals}). The estimates of the classifier are on average 6.6 points from the true racial gap on the geographic outcomes and 24.5 points on party registration. Estimates also differ across BISG, BIRDiE, and the LLM, usually by much more than the standard errors. A low calibration error similarly does not ensure unbiased downstream estimates. On the geographic racial gaps, the probability-weighted estimate from the distilled classifier is further from the truth (6.6 points on average) than the LLM from which the classifier is distilled (2.1 points). 

\subsection{Neighborhood Misclassification Gradient}
\label{app:gradient}

\begin{figure}[!t]
\centering
\includegraphics[width=.85\linewidth]{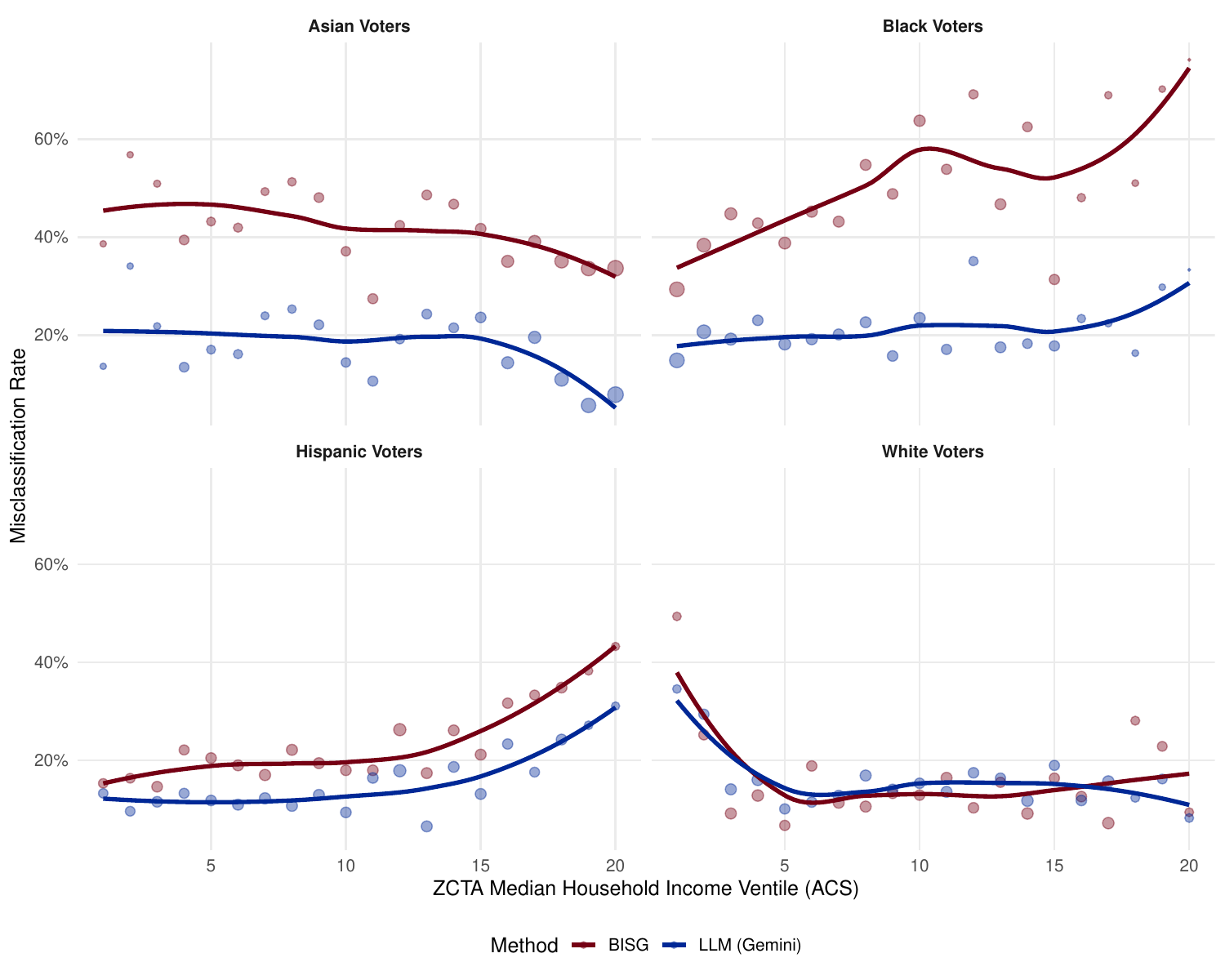}
\caption{Misclassification rate by ZCTA median household income and race, North Carolina voters. Points are averages within income ventiles, sized by the number of voters, and lines are loess fits.}
\label{fig:income_bias}
\end{figure}

To test whether the income gradient plotted for North Carolina in Figure~\ref{fig:income_bias} holds more generally, I extend the analysis to all of the neighborhood measures tested in \citet{argyle2024misclassification}, three different minority groups, four states, and two geographic scales (Table~\ref{tab:gradient}). Among voters of each true race, I regress an indicator for being classified as White on a single neighborhood measure (income, education, or political activity), measured both at the ZCTA and the census tract of the voter (the same geographic units as \citet{argyle2024misclassification}). A positive coefficient matches their finding that minorities living in wealthier, more educated, or more politically active neighborhoods are more likely to be classified as White. 

The income gradient (i.e., relationship between misclassification rate and wealth of a neighborhood) is strongest for Black voters (Panel A of Table~\ref{tab:gradient}). Because Black voters' surnames overlap heavily with those of White voters, BISG is more reliant on geography, inducing geographic error in this way. Inferring race with LLMs instead of BISG (specifically Gemini~3.0 Flash as tested) about halves the income gradient for Black voters in North Carolina, Alabama, and South Carolina. More broadly, the BISG income and education gradients for Black voters are positive at both geographic levels, and outside Florida the LLM gradients are about half as strong on average. For Hispanic voters, the LLM gradient is sometimes stronger than that of BISG; however, Hispanic names that more closely resemble White ones are actually concentrated in wealthier neighborhoods. For Asian voters (Panel C), names are sufficiently distinctive from other groups such that BISG rarely relies on the geographic prior.

\begin{table}[!t]
\centering
\footnotesize
\begin{adjustbox}{max width=\textwidth}
\begin{tabular}{llcccc}
\toprule
 & & \multicolumn{2}{c}{BISG} & \multicolumn{2}{c}{LLM} \\
\cmidrule(lr){3-4}\cmidrule(lr){5-6}
State & Neighborhood covariate & ZCTA & Tract & ZCTA & Tract \\
\midrule
\multicolumn{6}{l}{\emph{Panel A: Black voters}} \\
Florida & Median income & $+0.088^{***}$ & $+0.026$ & $+0.102^{***}$ & $+0.083^{***}$ \\
 & Share with bachelor's & $+0.084^{**}$ & $+0.075^{**}$ & $+0.126^{***}$ & $+0.141^{***}$ \\
 & Political activity & $+0.413^{***}$ & $+0.144^{***}$ & $+0.340^{***}$ & $+0.248^{***}$ \\
\hdashline
North Carolina & Median income & $+0.167^{***}$ & $+0.078^{***}$ & $+0.066^{***}$ & $+0.028$ \\
 & Share with bachelor's & $+0.104^{***}$ & $+0.061^{**}$ & $+0.044$ & $+0.027$ \\
 & Political activity & $+0.223^{***}$ & $+0.046$ & $+0.139$ & $+0.032$ \\
\hdashline
Alabama & Median income & $+0.198^{***}$ & $+0.128^{***}$ & $+0.084^{***}$ & $+0.094^{***}$ \\
 & Share with bachelor's & $+0.129^{***}$ & $+0.062^{*}$ & $+0.049$ & $+0.051$ \\
 & Political activity & $-0.238^{***}$ & $-0.193^{***}$ & $-0.277^{***}$ & $-0.139^{**}$ \\
\hdashline
South Carolina & Median income & $+0.266^{***}$ & $+0.169^{***}$ & $+0.144^{***}$ & $+0.139^{***}$ \\
 & Share with bachelor's & $+0.184^{***}$ & $+0.150^{***}$ & $+0.076^{**}$ & $+0.117^{***}$ \\
 & Political activity & $-0.473^{***}$ & $-0.498^{***}$ & $-0.261^{***}$ & $-0.177^{**}$ \\
\midrule
\multicolumn{6}{l}{\emph{Panel B: Hispanic voters}} \\
Florida & Median income & $+0.106^{***}$ & $+0.073^{***}$ & $+0.087^{**}$ & $+0.082^{***}$ \\
 & Share with bachelor's & $+0.166^{***}$ & $+0.123^{**}$ & $+0.195^{***}$ & $+0.189^{***}$ \\
 & Political activity & $+0.959^{***}$ & $+0.784^{***}$ & $+0.828^{***}$ & $+0.664^{***}$ \\
\hdashline
North Carolina & Median income & $+0.166^{***}$ & $+0.120^{***}$ & $+0.167^{***}$ & $+0.162^{***}$ \\
 & Share with bachelor's & $+0.206^{***}$ & $+0.213^{***}$ & $+0.230^{***}$ & $+0.272^{***}$ \\
 & Political activity & $+0.557^{***}$ & $+0.488^{***}$ & $+0.432^{***}$ & $+0.529^{***}$ \\
\hdashline
Alabama & Median income & $+0.123^{***}$ & $+0.101^{***}$ & $+0.142^{***}$ & $+0.112^{***}$ \\
 & Share with bachelor's & $+0.137^{***}$ & $+0.136^{***}$ & $+0.189^{***}$ & $+0.155^{***}$ \\
 & Political activity & $+0.478^{***}$ & $+0.441^{***}$ & $+0.569^{***}$ & $+0.480^{***}$ \\
\hdashline
South Carolina & Median income & $+0.117^{***}$ & $+0.101^{***}$ & $+0.101^{***}$ & $+0.082^{***}$ \\
 & Share with bachelor's & $+0.160^{***}$ & $+0.165^{***}$ & $+0.143^{***}$ & $+0.125^{***}$ \\
 & Political activity & $+0.265^{***}$ & $+0.276^{***}$ & $+0.349^{***}$ & $+0.302^{***}$ \\
\midrule
\multicolumn{6}{l}{\emph{Panel C: Asian voters}} \\
Florida & Median income & $-0.109^{***}$ & $-0.062^{***}$ & $-0.037$ & $-0.015$ \\
 & Share with bachelor's & $-0.224^{***}$ & $-0.095^{***}$ & $-0.130^{***}$ & $-0.038$ \\
 & Political activity & $+0.331^{***}$ & $+0.288^{***}$ & $+0.308^{***}$ & $+0.240^{***}$ \\
\hdashline
North Carolina & Median income & $+0.025^{*}$ & $+0.014$ & $-0.073^{***}$ & $-0.066^{***}$ \\
 & Share with bachelor's & $-0.022$ & $+0.009$ & $-0.149^{***}$ & $-0.076^{*}$ \\
 & Political activity & $+0.113$ & $+0.050$ & $+0.163$ & $+0.121$ \\
\hdashline
Alabama & Median income & $-0.045^{***}$ & $-0.035^{**}$ & $-0.059^{***}$ & $-0.037^{*}$ \\
 & Share with bachelor's & $-0.172^{***}$ & $-0.132^{***}$ & $-0.155^{***}$ & $-0.099^{***}$ \\
 & Political activity & $-0.278^{***}$ & $-0.164^{***}$ & $-0.195^{**}$ & $-0.150^{**}$ \\
\hdashline
South Carolina & Median income & $+0.050^{**}$ & $-0.002$ & $-0.024$ & $-0.030$ \\
 & Share with bachelor's & $-0.043$ & $-0.116^{***}$ & $-0.123^{***}$ & $-0.109^{***}$ \\
 & Political activity & $-0.119$ & $-0.201^{***}$ & $-0.090$ & $-0.076$ \\
\bottomrule
\end{tabular}
\end{adjustbox}
\caption{Coefficients (log-odds) from logistic regressions of being classified as White on one neighborhood measure, among voters of each true race, at the ZCTA and census-tract levels. Income is per \$10{,}000, education per 10 percentage points of the share with a bachelor's degree, and political activity per 10 percentage points of mean neighborhood turnout in the 2012--2018 general elections. $^{***}p<0.01$, $^{**}p<0.05$, $^{*}p<0.10$.}
\label{tab:gradient}
\end{table}

For Asian voters, the income and education gradients are near zero or negative. The coefficient for turnout is positive for Hispanic voters in every state. For Black voters, it is positive in Florida and North Carolina and negative in Alabama and South Carolina. In the latter two states, Black neighborhoods with high turnout are mostly racially homogeneous, and BISG accurately classifies the race of these local residents.  

\begin{table}[!t]
\centering
\footnotesize
\begin{adjustbox}{max width=\textwidth}
\begin{tabular}{lccc}
\toprule
Model & Income & Education & Downstream \\
 & gradient & gradient & MAE (pp) \\
\midrule
\multicolumn{4}{l}{\textit{Proprietary (API-only)}} \\
Gemini~3.0 Flash & $0.084$ & $0.085$ & $2.1$ \\
GPT-4o & $0.042$ & $0.053$ & $2.7$ \\
GLM-5-Turbo & $0.077$ & $0.073$ & $2.9$ \\
Qwen3.5-Plus & $0.049$ & $0.040$ & $2.9$ \\
GPT-4.1 mini & $0.058$ & $0.085$ & $3.8$ \\
\hdashline
\multicolumn{4}{l}{\textit{Open-weight (downloadable, version-pinnable)}} \\
DeepSeek-V4 & $0.056$ & $0.043$ & $2.8$ \\
GLM-4.7 & $0.070$ & $0.053$ & $3.2$ \\
\midrule
\textit{BISG (surname + geography)} & $0.127$ & $0.094$ & $5.6$ \\
\bottomrule
\end{tabular}
\end{adjustbox}
\caption{Misclassification across models, Florida and North Carolina. The income and education gradients are the mean absolute ZCTA coefficients among true Black voters. Downstream MAE is the mean absolute error (percentage points) of the twelve racial gaps of Figure~\ref{fig:downstream}. GLM-4.7 is an earlier GLM model used only in this robustness check.}
\label{tab:gradient_models}
\end{table}

To confirm that these patterns are not dependent on a single model, I replicate both analyses for seven different LLMs, five of which are proprietary and two open-weight, on both the Florida and North Carolina samples (Table~\ref{tab:gradient_models}). Every model, including the open-weight ones, flattens the income and education gradients for Black voters relative to BISG. Each model also estimates the racial gaps of Figure~\ref{fig:downstream} more accurately than BISG.

\subsection{BIRDiE Correction}
\label{app:birdie_llm}

BIRDiE \citep{mccartan2025birdie}, the standard correction for measurement error in the setting of Bayesian surname geocoding, accepts race probabilities as inputs and makes additional assumptions about the outcome and predictors. On the geographic outcomes considered here, BIRDiE overestimates the racial gap that the LLM is able to recover closely (Figure~\ref{fig:downstream}). 

I also test BIRDiE with the probabilities of the distilled logistic classifier from Section~\ref{subsec:uncertainty}, which has a lower calibration error than the raw LLM probabilities in these samples. I use the same two states and three outcomes as in Figure~\ref{fig:downstream}. The geographic outcomes are frequent turnout (voting in at least half of the general elections from 2012 to 2018) and living in an above-median-income ZCTA. The third outcome, being a registered Democrat, has a large racial gap that is not primarily geographic in nature. In Table~\ref{tab:birdie_llm}, I add two variants that apply BIRDiE to the LLM probabilities. I report the mean absolute error of the estimated racial gap between minority and White voters for each method.

\begin{table}[!t]
\centering
\small
\begin{adjustbox}{max width=\textwidth}
\begin{tabular}{lcc c}
\toprule
 & \multicolumn{2}{c}{Geographic} & Non-geographic \\
\cmidrule(lr){2-3}\cmidrule(lr){4-4}
Method & Turnout & Higher-income & Reg.\ Democrat \\
\midrule
BISG (naive) & 4.1 & 7.2 & 22.7 \\
BISG + BIRDiE & 3.1 & 7.0 & 5.4 \\
LLM (name + geo), categorical & \textbf{1.9} & \textbf{2.4} & 10.8 \\
LLM classifier (name) + BIRDiE & 7.4 & 6.2 & \textbf{4.2} \\
LLM classifier (name + geo) + BIRDiE & 8.7 & 9.7 & 7.7 \\
\bottomrule
\end{tabular}
\end{adjustbox}
\caption[BIRDiE correction]{Mean absolute error (percentage points) of the estimated gap between minority and White voters, averaged over three groups and two states. The ``$+$ BIRDiE'' rows apply the \texttt{cat\_dir} correction of \citet{mccartan2025birdie} to BISG and to the probabilities of the LLM logistic classifier. Bold indicates the most accurate method for each outcome. Estimates for each group and state are in Table~\ref{tab:birdie_by_group}.}
\label{tab:birdie_llm}
\end{table}

\begin{table}[!t]
\centering
\footnotesize
\begin{adjustbox}{max width=\textwidth}
\begin{tabular}{l|ccc|ccc|ccc}
\toprule
 & \multicolumn{3}{c|}{Turnout} & \multicolumn{3}{c|}{Higher-income} & \multicolumn{3}{c}{Reg.\ Democrat} \\
Method & Black & Hispanic & Asian & Black & Hispanic & Asian & Black & Hispanic & Asian \\
\midrule
\multicolumn{10}{l}{\emph{Panel A: Florida}} \\
\textit{True gap} & $-$6.5 & $-$13.1 & $-$17.6 & $-$28.6 & $-$3.5 & 7.5 & 53.1 & 10.5 & 6.0 \\
\hdashline
BISG (naive) & $-$1.8 & $-$7.7 & $-$16.7 & $-$12.4 & 3.2 & 15.0 & 20.8 & $-$4.1 & $-$8.9 \\
BISG + BIRDiE & $-$5.2 & $-$11.6 & $-$23.7 & $-$40.0 & $-$0.8 & 13.0 & 62.9 & 6.1 & $-$2.1 \\
LLM (name + geo), categorical & $-$6.4 & $-$11.5 & $-$18.9 & $-$24.1 & $-$1.0 & 11.8 & 38.9 & 0.8 & $-$1.5 \\
LLM classifier (name) + BIRDiE & $-$16.1 & $-$16.5 & $-$27.3 & $-$38.1 & $-$4.3 & 12.9 & 61.8 & 5.9 & 2.1 \\
LLM classifier (name + geo) + BIRDiE & $-$15.7 & $-$17.2 & $-$30.1 & $-$45.9 & $-$5.2 & 12.6 & 71.3 & 7.4 & 3.4 \\
\midrule
\multicolumn{10}{l}{\emph{Panel B: North Carolina}} \\
\textit{True gap} & $-$2.3 & $-$29.2 & $-$25.9 & $-$15.8 & 0.6 & 20.1 & 58.3 & 17.9 & 6.5 \\
\hdashline
BISG (naive) & 1.2 & $-$23.5 & $-$21.4 & $-$7.5 & 1.0 & 24.0 & 17.4 & 1.0 & $-$10.5 \\
BISG + BIRDiE & 2.3 & $-$32.4 & $-$27.6 & $-$26.7 & $-$6.9 & 24.5 & 61.1 & 16.0 & 0.7 \\
LLM (name + geo), categorical & 0.4 & $-$26.6 & $-$22.8 & $-$14.9 & $-$1.4 & 20.2 & 40.5 & 10.4 & $-$1.7 \\
LLM classifier (name) + BIRDiE & $-$6.5 & $-$38.2 & $-$34.6 & $-$29.6 & $-$5.4 & 22.1 & 63.8 & 18.0 & 4.4 \\
LLM classifier (name + geo) + BIRDiE & $-$6.1 & $-$41.1 & $-$36.6 & $-$36.7 & $-$4.9 & 27.5 & 75.4 & 22.0 & 7.7 \\
\bottomrule
\end{tabular}
\end{adjustbox}
\caption{Estimated gap (percentage points) between each minority group and White voters. The first row of each panel is the gap measured with true race.}
\label{tab:birdie_by_group}
\end{table}

With respect to the two geographic outcomes, predictions from the categorical LLM produce the most accurate downstream estimates of the racial gap, while the BIRDiE variants misestimate the gap, mostly by overstating it. On party registration, for which BIRDiE was specifically developed with the outcome in mind, BIRDiE is more accurate than the LLM. 

Applied to the probabilities from the name-only logistic classifier, BIRDiE corrects the understated party racial gap and is the most accurate method on that outcome, more accurate even than BIRDiE applied to BISG. On the geographic outcomes, however, this combination also overestimates the racial gaps. Adding geography to the classifier increases the extent of overestimation on every outcome. This overestimate does not depend on the settings of BIRDiE, which I run under the default configuration as well as different priors (uninformative, strongly regularizing). BIRDiE assumes that the outcome is independent of the probability inputs given race, which fails for an outcome that itself strongly varies with geography. 

\subsection{Agreement Between Models}
\label{app:agreement}

The accuracy of ethnicity inference with LLMs depends on the extent to which names are representative of a given target group. However, a researcher who applies the method to a new population has no true labels with which to verify this in advance. Agreement between models provides one option for checking whether this is the case, and does not require ground truth labels. In each setting, I classify the same individuals with three different models (DeepSeek-V4, GLM-5-Turbo, and Qwen3.5-Plus), and measure how often the three models agree. 

\begin{table}[!t]
\centering
\small
\begin{adjustbox}{max width=\textwidth}
\begin{tabular}{lcc}
\toprule
Setting & Inter-model agreement (\%) & Accuracy (\%) \\
\midrule
Lee and Velez (race) & 90.4 & 86.4 \\
North Carolina (race) & 87.2 & 78.4 \\
Florida (race) & 86.4 & 75.2 \\
Alabama (race) & 85.7 & 73.4 \\
South Carolina (race) & 84.9 & 70.8 \\
Lebanon (sect) & 58.5 & 48.3 \\
India, Bihar (caste) & 46.6 & 58.2 \\
India, Rajasthan (caste) & 30.7 & 35.0 \\
\bottomrule
\end{tabular}
\end{adjustbox}
\caption{Agreement between models and accuracy, by setting. Agreement is the mean rate at which each pair of DeepSeek-V4, GLM-5-Turbo, and Qwen3.5-Plus assigns the same category to the same person. Accuracy is the mean accuracy of the three models on the records that are classified by all three.}
\label{tab:agreement}
\end{table}

Across the eight settings, this agreement and the accuracy of those three models have a Spearman rank correlation of 0.98 and a Pearson correlation of 0.94 (Table~\ref{tab:agreement}). Where the models agree, they are usually correct, and where they disagree, accuracy tends to be lower. A researcher can thus run two or three models on an unlabeled sample and use the agreement between them as an early indication of whether the task is feasible. This procedure may also help determine \emph{which} groups to classify individuals into; for instance, where a given ethnic group has different sub-groups, agreement may indicate which of those sub-groups, if any, models can plausibly distinguish. This is far less costly than hand-coding a classification set.

\subsection{Memorization}
\label{app:access}

One potential concern with the use of LLMs for ethnicity inference is that they are simply memorizing known name-ethnicity relationships in the training data, and thus perform poorly when the data are not in their existing corpora. However, several results are inconsistent with the premise of the models simply recalling voters' races from their training data. First, the accuracy of LLMs holds for voters who registered to vote after the training data cut-off (below), as well as for synthetic names that I generate which are not found in any voter file (Appendix~\ref{app:synthetic}). Additionally, the advantage of LLMs over BISG is largest for surnames that are missing from the Census surname table (Appendix~\ref{app:rarity}). 

I also demonstrate that the advantage does not depend on how easily the given voter file of a state can be obtained. North Carolina, for instance, publishes the state voter file as a free download at regular intervals, and thus it is more likely that names and races appear in the training data. Florida releases the voter file only upon formal request. I also classify voters in two states whose files have never been posted online: Alabama, where the statewide voter file costs about \$37,000 (the highest price in the country); and South Carolina, which restricts bulk access to the state voter file.\footnote{Pricing and access information from the U.S.\ Election Assistance Commission, \emph{Availability of State Voter File and Confidential Information} (2020). Alabama and South Carolina also record race as a mandatory registration field.} If memorization explained the strong performance for LLMs, the advantage over BISG would be confined to the freely available file(s), or at the least scale with the availability of different state files. Instead, the advantage of LLMs over BISG holds across all four states (Table~\ref{tab:durability}).

Two further points support this conclusion. First, even a model that had ``seen'' voter records in training would still need to apply the association between names and race to the individuals in the sample. Second, memorization is implausible in several of the international cases, especially where there are no ground truth labels. The Lebanese voter file is not available online, and I collected the full voter files with no ground truth ethnic labels for Uganda, Nepal, Armenia, Chile, and Costa Rica. As in the United States, accuracy in these cases is higher where names are more distinctive to the given ethnic group.

\phantomsection
\label{app:postcutoff}

One stronger test varies exposure for each person. A person who first registered to vote after the training data of a model were collected cannot by nature appear in those data. Assuming memorization and recall were the bases for strong performance by LLMs, the model would classify these voters less accurately than voters registered long before the training cutoff. I draw two groups from the September 2025 North Carolina file, each with 100 voters per racial category. The first group registered during 2025, and the second registered in 2010 or earlier. I classify both groups with Gemini~3.0 Flash at low reasoning, using full name and city of residence. The model accuracy is 81.8\% for the voters who first registered in 2025, and 84.5\% for those who registered by 2010. At this sample size, the difference is within sampling error, and is much smaller than the difference that simple memorization would suggest. 

\subsection{Synthetic Names}
\label{app:synthetic}

\begin{table}[!t]
\centering
\small
\begin{adjustbox}{max width=\textwidth}
\begin{tabular}{lcccc}
\toprule
& \multicolumn{2}{c}{North Carolina} & \multicolumn{2}{c}{Florida} \\
\cmidrule(lr){2-3}\cmidrule(lr){4-5}
Racial category & Real & Synthetic & Real & Synthetic \\
\midrule
White & 89.6 & 90.2 & 91.4 & 91.0 \\
Black & 74.0 & 76.0 & 77.4 & 77.2 \\
Hispanic & 81.2 & 84.4 & 93.2 & 93.8 \\
Asian & 83.0 & 87.2 & 70.8 & 74.8 \\
\midrule
All & 82.0 & 84.5 & 83.2 & 84.2 \\
\bottomrule
\end{tabular}
\end{adjustbox}
\caption{Accuracy (\%) of Gemini~3.0 Flash on real and synthetic names. Each synthetic name pairs a real first-and-middle name and a real surname from the same category.}
\label{tab:synthetic}
\end{table}

A second test against memorization uses names that are not ``real,'' in that they do not belong to any registered voters that we can observe. Within each racial category, I randomly shuffle the first and middle names and the surnames of the sampled voters, retaining the county of each record, then construct synthetic full names from these shuffles. Every synthetic voter therefore has a real first name and a real surname from the same racial category, but the full name does not appear in any of the voter files. If accuracy were the result of simple recall, it would fall on these synthetic names. Instead, on balanced samples of 2{,}000 voters in each of North Carolina and Florida, Gemini~3.0 Flash classifies the synthetic names at least as accurately as the real names. Accuracy across categories falls by no more than half a point (Table~\ref{tab:synthetic}). Of course, it should also be acknowledged that these synthetic names may enter the training data through other sources beyond the observed voter files.

\subsection{Surname Frequency}
\label{app:rarity}

\begin{table}[!t]
\centering
\small
\begin{adjustbox}{max width=\textwidth}
\begin{tabular}{lrccc}
\toprule
Surname frequency (2010 Census) & n & Gemini & BISG & Gemini $-$ BISG \\
\midrule
Not in the Census list & 2{,}288 & 86.5 & 42.9 & $+43.6$ \\
Rare ($<$1{,}000) & 2{,}091 & 85.1 & 76.9 & $+8.2$ \\
Uncommon (1{,}000--20{,}000) & 5{,}305 & 83.9 & 73.7 & $+10.2$ \\
Common ($\geq$20{,}000) & 10{,}316 & 82.6 & 69.9 & $+12.7$ \\
\midrule
All & 20{,}000 & 83.6 & 68.6 & $+15.1$ \\
\bottomrule
\end{tabular}
\end{adjustbox}
\caption{Accuracy (\%) by the frequency of the surname in the 2010 Census surname list, Florida and North Carolina pooled.}
\label{tab:rarity}
\end{table}

If the models simply recalled memorized or well-known names instead of generalizing from the semantic form of a given name, the advantage of LLMs over BISG would be greatest on common surnames and smallest on rare names. To test this, I pool the Florida and North Carolina samples and group records together based on how often the surname appears in the 2010 Census surname list (Table~\ref{tab:rarity}). I find that BISG is much less accurate than LLMs for surnames that are wholly absent from the list, because BISG relies on the surname distribution and without it infers solely on the basis of the geographic prior. Conversely, Gemini~3.0 Flash classifies these names as accurately as it does others. The advantage of LLMs over BISG is therefore largest, not smallest, for surnames absent from the Census list, which are the least likely to be memorized and recalled. Among surnames in the list, the advantage is larger for common surnames than for rare ones, but it holds in every surname frequency band. 

\section{Agentic Validation}
\label{app:agentic}

In the agentic audit of the held-out ParaNames matches (Table~\ref{tab:agentic}), the agent mostly declines to form a final verdict for obscure entities and for names that are shared by several people or places, which no method is able to reliably link with the information that is provided.

\begin{table}[!t]
\centering
\small
\begin{adjustbox}{max width=\textwidth}
\begin{tabular}{lcc}
\toprule
 & Record Linkage & Ethnicity \\
 & (ParaNames, held-out) & (Lee \& Velez) \\
\midrule
Sample size & 100 pairs & 108 individuals \\
Confirmed / Agreed & 49 (49.0\%) & 103 (98.1\% of 105) \\
Rejected / Disagreed & 20 (20.0\%) & 2 (1.9\% of 105) \\
Uncertain & 31 & 3 \\
Agent agrees with gold label & 64 (92.8\% of 69) & --- \\
Part-whole & 1 & --- \\
Gold-label issues flagged & 1 & 4 (3.7\%) \\
Identical verdicts across two runs & 89\% & --- \\
Validation time per run & about 20 minutes & minutes \\
\bottomrule
\end{tabular}
\end{adjustbox}
\caption{Agentic validation of 100 held-out ParaNames matches from EnsembleLink, half drawn from the lowest quarter of EnsembleLink confidence, and of 108 Lee and Velez officials. The agent is provided the original name, the entity type, and the matched English name, but not the gold label. Agreement with the gold label is computed over the pairs that the agent judges. ``Part-whole'' counts rejected pairs in which one entity is part of the other. Record-linkage values are from the first of two runs.}
\label{tab:agentic}
\end{table}

Among the pairs that the agent does judge, the few disagreements with the gold labels mostly involve namesakes, such as two villages with the same name for which the agent should have withheld a verdict. In one flagged pair, the benchmark ``ground truth'' is arguably wrong. Specifically, the gold entity is a women's national cycling team, but the benchmark entry omits ``women's'' from the name. EnsembleLink returned the national team rather than the women's national team, a match that is arguably correct. 

\clearpage

%% file: tables/tab_linkage_summary.tex
\begin{tabular}{lccccc}
\toprule
Benchmark & EnsembleLink & fuzzylink (validation data) & fuzzylink (test data) & fastLink & Splink \\
\midrule
NC voters, address changes & \textbf{97.5} & 96.6 & 97.1 & 96.7 & 97.4 \\
NC voters, address changes, 20\% unmatched & 96.1 & 94.3 & \textbf{96.9} & 95.6 & 95.3 \\
MaDI pairs, five domains & 86.0 & 87.0 & \textbf{91.5} & 78.4 & 81.5 \\
MaDI pairs, five domains, hard variants & 83.2 & 77.9 & \textbf{87.8} & 64.8 & 66.3 \\
MaDI products, full table join & 67.1 & 59.6 & \textbf{70.4} & 40.9 & 38.2 \\
ParaNames, six languages & \textbf{75.9} & 49.3 & 66.6 & 5.5 & 3.7 \\
GeoNames, name only & \textbf{70.1} & 28.8 & 50.7 & 8.5 & 7.9 \\
GeoNames, name with country and region & \textbf{95.1} & 83.7 & 93.1 & 32.3 & 31.8 \\
Census occupations to OCC1950 titles & 53.9 & 40.1 & \textbf{65.9} & 26.2 & 26.2 \\
Bill summaries to bill texts & 94.8 & 95.3 & \textbf{95.5} & 0.1 & 0.1 \\
Wikipedia articles, six languages to English & \textbf{99.5} & 98.7 & 99.4 & 0.0 & 0.0 \\
\bottomrule
\end{tabular}

%% file: tables/tab_nc_detail.tex
\begin{tabular}{lrrrr}
\toprule
Method & Precision & Recall & F1 & False links \\
\midrule
\multicolumn{5}{l}{\emph{All queries matched}} \\
EnsembleLink & 98.9 & 96.1 & \textbf{97.5} & --- \\
fuzzylink & 99.8 & 94.5 & 97.1 & --- \\
fastLink & 97.3 & 96.2 & 96.7 & --- \\
Splink & 98.4 & 96.4 & 97.4 & --- \\
Jaro-Winkler & 99.1 & 94.8 & 96.9 & --- \\
\multicolumn{5}{l}{\emph{20\% unmatched}} \\
EnsembleLink & 97.9 & 94.4 & 96.1 & 6.5 \\
fuzzylink & 99.6 & 94.4 & \textbf{96.9} & 1.0 \\
fastLink & 99.2 & 92.2 & 95.6 & 1.8 \\
Splink & 97.3 & 93.4 & 95.3 & 8.8 \\
Jaro-Winkler & 98.0 & 93.9 & 95.9 & 4.5 \\
\multicolumn{5}{l}{\emph{Voters whose address did not change (diagnostic)}} \\
EnsembleLink & 99.4 & 99.2 & 99.3 & --- \\
fuzzylink & 99.8 & 99.0 & \textbf{99.4} & --- \\
fastLink & 99.0 & 98.8 & 98.9 & --- \\
Splink & 99.2 & 98.2 & 98.7 & --- \\
Jaro-Winkler & 98.2 & 97.6 & 97.9 & --- \\
\bottomrule
\end{tabular}

%% file: tables/tab_madi_detail.tex
\begin{tabular}{lrrrrr}
\toprule
Method & Companies & Games & Music & Papers & Products \\
\midrule
\multicolumn{6}{l}{\emph{Unaltered records}} \\
EnsembleLink & 87.6 & 69.6 & \textbf{96.7} & \textbf{99.8} & 76.4 \\
fuzzylink & \textbf{94.9} & \textbf{76.3} & 95.4 & 99.2 & \textbf{91.7} \\
fastLink & 77.1 & 49.3 & 93.9 & 99.4 & 72.5 \\
Splink & 81.1 & 57.1 & 92.6 & 99.6 & 76.9 \\
Jaro-Winkler & 81.8 & 64.1 & 90.3 & 99.1 & 73.4 \\
\multicolumn{6}{l}{\emph{Easy version}} \\
EnsembleLink & 88.2 & 70.6 & 95.2 & \textbf{99.6} & 88.7 \\
fuzzylink & \textbf{94.6} & \textbf{71.5} & \textbf{96.0} & 98.8 & \textbf{92.9} \\
fastLink & 78.5 & 40.4 & 93.0 & 97.1 & 86.4 \\
Splink & 81.3 & 61.0 & 92.8 & 97.5 & 84.8 \\
Jaro-Winkler & 83.5 & 49.8 & 91.9 & 98.6 & 86.4 \\
\multicolumn{6}{l}{\emph{Medium version}} \\
EnsembleLink & 88.0 & 67.4 & 93.2 & \textbf{99.3} & 85.2 \\
fuzzylink & \textbf{95.2} & \textbf{71.2} & \textbf{94.3} & 98.3 & \textbf{92.9} \\
fastLink & 74.9 & 0.0 & 89.3 & 97.1 & 75.3 \\
Splink & 81.9 & 18.2 & 89.3 & 97.4 & 67.2 \\
Jaro-Winkler & 77.9 & 3.4 & 86.1 & 98.0 & 69.6 \\
\multicolumn{6}{l}{\emph{Hard version}} \\
EnsembleLink & 85.9 & 64.0 & \textbf{92.5} & \textbf{98.1} & 75.4 \\
fuzzylink & \textbf{95.5} & \textbf{66.4} & 91.2 & 96.7 & \textbf{89.3} \\
fastLink & 65.0 & 15.6 & 84.0 & 95.3 & 64.0 \\
Splink & 78.7 & 23.6 & 83.7 & 95.4 & 50.3 \\
Jaro-Winkler & 70.2 & 28.9 & 78.7 & 95.7 & 66.4 \\
\bottomrule
\end{tabular}

%% file: tables/tab_products_detail.tex
\begin{tabular}{lrrrr}
\toprule
Method & Precision & Recall & F1 & False links \\
\midrule
\multicolumn{5}{l}{\emph{products 1 to products 2}} \\
EnsembleLink & 71.7 & 65.6 & 68.5 & --- \\
fuzzylink & 89.1 & 56.6 & \textbf{69.2} & --- \\
fastLink & 45.2 & 37.5 & 41.0 & --- \\
Splink & 44.1 & 37.3 & 40.4 & --- \\
Jaro-Winkler & 50.5 & 50.3 & 50.4 & --- \\
\multicolumn{5}{l}{\emph{products 1 to products 3}} \\
EnsembleLink & 68.0 & 65.9 & 67.0 & 73.3 \\
fuzzylink & 88.5 & 56.8 & \textbf{69.2} & 3.3 \\
fastLink & 43.3 & 39.7 & 41.4 & 80.0 \\
Splink & 38.4 & 34.9 & 36.6 & 60.0 \\
Jaro-Winkler & 49.5 & 51.6 & 50.5 & 83.3 \\
\multicolumn{5}{l}{\emph{products 1 to products 4}} \\
EnsembleLink & 66.6 & 65.1 & 65.8 & 52.9 \\
fuzzylink & 91.4 & 60.5 & \textbf{72.8} & 8.3 \\
fastLink & 41.4 & 39.5 & 40.4 & 57.9 \\
Splink & 37.3 & 38.1 & 37.7 & 55.4 \\
Jaro-Winkler & 41.5 & 50.0 & 45.3 & 76.0 \\
\bottomrule
\end{tabular}

%% file: tables/tab_names_detail.tex
\begin{tabular}{lrrrr}
\toprule
Method & Precision & Recall & F1 & False links \\
\midrule
\multicolumn{5}{l}{\emph{ParaNames, all queries matched}} \\
EnsembleLink & 85.0 & 69.5 & \textbf{75.9} & --- \\
fuzzylink & 78.1 & 58.8 & 66.6 & --- \\
fastLink & 8.9 & 4.2 & 5.5 & --- \\
Splink & 6.7 & 2.6 & 3.7 & --- \\
Transliteration and Jaro-Winkler & 45.4 & 29.7 & 35.0 & --- \\
\multicolumn{5}{l}{\emph{ParaNames, 20\% unmatched}} \\
EnsembleLink & 80.4 & 66.4 & \textbf{72.4} & 33.6 \\
fuzzylink & 72.8 & 57.7 & 63.8 & 28.9 \\
fastLink & 7.7 & 4.2 & 5.2 & 88.4 \\
Splink & 4.9 & 2.4 & 3.1 & 88.2 \\
Transliteration and Jaro-Winkler & 40.7 & 28.0 & 32.6 & 42.6 \\
\multicolumn{5}{l}{\emph{GeoNames, all queries matched}} \\
EnsembleLink & 88.8 & 78.1 & \textbf{82.6} & --- \\
fuzzylink & 84.4 & 65.5 & 71.9 & --- \\
fastLink & 22.4 & 18.9 & 20.4 & --- \\
Splink & 22.3 & 18.2 & 19.9 & --- \\
Transliteration and Jaro-Winkler & 72.0 & 62.8 & 66.2 & --- \\
\multicolumn{5}{l}{\emph{GeoNames, 20\% unmatched}} \\
EnsembleLink & 84.0 & 76.1 & \textbf{79.5} & 28.8 \\
fuzzylink & 81.4 & 65.2 & 70.8 & 14.0 \\
fastLink & 19.7 & 18.9 & 19.0 & 82.5 \\
Splink & 19.6 & 18.1 & 18.5 & 82.0 \\
Transliteration and Jaro-Winkler & 66.7 & 58.6 & 61.7 & 41.4 \\
\bottomrule
\end{tabular}

%% file: tables/tab_occupations_detail.tex
\begin{tabular}{lrrrrrr}
\toprule
Method & Top-1 & Top-1, near title & Top-1, other & Precision & Recall & F1 \\
\midrule
EnsembleLink & 52.3 & 84.9 & 37.7 & 61.0 & 48.3 & 53.9 \\
fuzzylink & 59.6 & 84.9 & 48.2 & 73.7 & 59.6 & \textbf{65.9} \\
fastLink & 16.5 & 45.5 & 3.5 & 67.9 & 16.2 & 26.2 \\
Splink & 16.5 & 45.5 & 3.5 & 67.8 & 16.2 & 26.2 \\
Jaro-Winkler & 30.2 & 45.5 & 23.3 & 46.4 & 29.0 & 35.7 \\
\bottomrule
\end{tabular}

%% file: tables/tab_longform_detail.tex
\begin{tabular}{lrrrr}
\toprule
Method & Top-1 & Precision & Recall & F1 \\
\midrule
\multicolumn{5}{l}{\emph{Bill summaries to bill texts, all queries matched}} \\
EnsembleLink & 94.8 & 95.3 & 94.4 & 94.8 \\
fuzzylink & 93.6 & 97.5 & 93.6 & \textbf{95.5} \\
fastLink & 0.1 & 0.1 & 0.1 & 0.1 \\
Splink & 0.1 & 0.1 & 0.1 & 0.1 \\
Jaro-Winkler & 0.9 & 2.2 & 0.8 & 1.2 \\
\multicolumn{5}{l}{\emph{Bill summaries to bill texts, 20\% unmatched}} \\
EnsembleLink & 94.6 & 91.8 & 88.1 & 89.9 \\
fuzzylink & 92.2 & 95.3 & 92.2 & \textbf{93.7} \\
fastLink & 0.1 & 0.1 & 0.1 & 0.1 \\
Splink & 0.1 & 0.1 & 0.1 & 0.1 \\
Jaro-Winkler & 1.0 & 2.1 & 0.4 & 0.7 \\
\multicolumn{5}{l}{\emph{Wikipedia articles, six languages to English, all queries matched}} \\
EnsembleLink & 99.5 & 99.7 & 99.4 & \textbf{99.5} \\
fuzzylink & 98.9 & 100.0 & 98.9 & 99.4 \\
fastLink & 0.0 & 0.0 & 0.0 & 0.0 \\
Splink & 0.0 & 0.0 & 0.0 & 0.0 \\
Jaro-Winkler & 8.1 & 67.0 & 5.1 & 9.5 \\
\multicolumn{5}{l}{\emph{Wikipedia articles, six languages to English, 20\% unmatched}} \\
EnsembleLink & 99.8 & 97.5 & 99.2 & 98.4 \\
fuzzylink & 98.9 & 99.9 & 98.9 & \textbf{99.4} \\
fastLink & 0.0 & 0.0 & 0.0 & 0.0 \\
Splink & 0.0 & 0.0 & 0.0 & 0.0 \\
Jaro-Winkler & 8.1 & 55.2 & 4.8 & 8.9 \\
\bottomrule
\end{tabular}

%% file: tables/tab_linkage_ranking.tex
\begin{tabular}{lrrrrr}
\toprule
Benchmark & EnsembleLink & fuzzylink (validation data) & fuzzylink (test data) & fastLink & Splink \\
\midrule
\multicolumn{6}{l}{\emph{Top-1 accuracy (\%)}} \\
\midrule
NC voters, address changes & \textbf{97.8} & 96.5 & 94.5 & 96.7 & 97.4 \\
NC voters, 20\% unmatched & \textbf{97.6} & 96.6 & 94.4 & 96.8 & 97.3 \\
ParaNames, six languages & \textbf{73.6} & 48.7 & 58.8 & 4.2 & 2.6 \\
ParaNames, six languages, 20\% unmatched & \textbf{74.0} & 48.7 & 57.7 & 4.2 & 2.4 \\
GeoNames, name only & \textbf{67.6} & 28.6 & 40.7 & 7.6 & 6.9 \\
GeoNames, name only, 20\% unmatched & \textbf{67.2} & 28.3 & 40.1 & 7.4 & 6.7 \\
GeoNames, name with country and region & \textbf{94.8} & 83.7 & 90.2 & 31.4 & 30.9 \\
GeoNames, name with country and region, 20\% unmatched & \textbf{95.1} & 83.9 & 90.3 & 31.4 & 30.9 \\
MaDI pairs, five domains & --- & --- & --- & --- & --- \\
MaDI pairs, hard versions & --- & --- & --- & --- & --- \\
MaDI products, full table join & \textbf{66.9} & 62.2 & 58.0 & 39.1 & 38.2 \\
Census occupations to OCC1950 titles & 52.3 & 37.5 & \textbf{59.6} & 16.5 & 16.5 \\
Bill summaries to bill texts & 94.8 & \textbf{95.3} & 93.6 & 0.1 & 0.1 \\
Bill summaries to bill texts, 20\% unmatched & 94.6 & \textbf{95.1} & 92.2 & 0.1 & 0.1 \\
Wikipedia articles, six languages to English & \textbf{99.5} & 98.8 & 98.9 & 0.0 & 0.0 \\
Wikipedia articles, six languages to English, 20\% unmatched & \textbf{99.8} & 98.8 & 98.9 & 0.0 & 0.0 \\
\midrule
\multicolumn{6}{l}{\emph{PR-AUC (average precision, \%)}} \\
\midrule
NC voters, address changes & \textbf{97.7} & 96.3 & 94.5 & 96.5 & 97.3 \\
NC voters, 20\% unmatched & \textbf{97.3} & 96.0 & 94.2 & 96.1 & 96.9 \\
ParaNames, six languages & \textbf{71.2} & 36.6 & 50.0 & 2.8 & 1.1 \\
ParaNames, six languages, 20\% unmatched & \textbf{69.0} & 33.3 & 46.7 & 2.5 & 0.7 \\
GeoNames, name only & \textbf{64.2} & 15.5 & 28.9 & 4.9 & 4.3 \\
GeoNames, name only, 20\% unmatched & \textbf{61.2} & 13.5 & 26.3 & 4.3 & 3.7 \\
GeoNames, name with country and region & \textbf{94.4} & 78.4 & 87.7 & 16.1 & 15.5 \\
GeoNames, name with country and region, 20\% unmatched & \textbf{93.6} & 75.6 & 86.8 & 14.6 & 14.0 \\
MaDI pairs, five domains & 90.0 & \textbf{93.0} & 88.0 & 79.1 & 83.7 \\
MaDI pairs, hard versions & 86.6 & \textbf{90.0} & 82.9 & 71.5 & 75.7 \\
MaDI products, full table join & \textbf{59.5} & 44.4 & 52.2 & 15.6 & 16.5 \\
Census occupations to OCC1950 titles & 39.6 & 24.5 & \textbf{48.4} & 13.1 & 13.1 \\
Bill summaries to bill texts & \textbf{94.1} & 92.9 & 91.6 & 0.0 & 0.0 \\
Bill summaries to bill texts, 20\% unmatched & \textbf{91.8} & 90.6 & 89.1 & 0.0 & 0.0 \\
Wikipedia articles, six languages to English & \textbf{99.5} & 98.7 & 98.9 & 0.0 & 0.0 \\
Wikipedia articles, six languages to English, 20\% unmatched & \textbf{99.5} & 98.3 & 98.9 & 0.0 & 0.0 \\
\bottomrule
\end{tabular}

%% file: tables/tab_jw_sensitivity.tex
\begin{tabular}{llrrrrrr}
\toprule
 & & \multicolumn{3}{c}{All queries matched} & \multicolumn{3}{c}{20\% unmatched} \\
\cmidrule(lr){3-5}\cmidrule(lr){6-8}
Method & Agreement cut-point & Precision & Recall & F1 & Precision & Recall & F1 \\
\midrule
fastLink & 0.94 (default) & 96.4 & 95.2 & \textbf{95.8} & 99.6 & 91.0 & \textbf{95.1} \\
fastLink & 0.90 & 95.9 & 94.7 & 95.3 & 97.7 & 91.1 & 94.3 \\
Splink & 0.94 (default) & 97.6 & 95.3 & \textbf{96.5} & 97.7 & 92.1 & \textbf{94.9} \\
Splink & 0.90 & 97.7 & 95.3 & \textbf{96.5} & 97.7 & 92.0 & 94.8 \\
\bottomrule
\end{tabular}

%% file: tables/tab_common_threshold.tex
\begin{tabular}{lccccccc}
\toprule
 & \multicolumn{5}{c}{EnsembleLink} & fastLink & Splink \\
\cmidrule(lr){2-6}\cmidrule(lr){7-8}
Benchmark & Task-specific & Leave-one-out & One cutoff & Reranker $\geq$ 0.5 & No cutoff & Probability $\geq$ 0.5 & Probability $\geq$ 0.5 \\
\midrule
NC voters, address changes & 97.5 & 92.3 & 92.3 & \textbf{97.8} & \textbf{97.8} & 96.0 & 97.2 \\
NC voters, address changes, 20\% unmatched & \textbf{96.1} & 95.6 & 95.6 & 89.2 & 86.7 & 95.1 & 93.5 \\
MaDI pairs, five domains & \textbf{86.0} & 84.6 & 85.5 & 85.5 & 44.3 & 78.8 & 73.8 \\
MaDI pairs, five domains, hard variants & 83.2 & 81.2 & 81.7 & \textbf{83.3} & 40.2 & 69.7 & 67.9 \\
MaDI products, full table join & \textbf{67.1} & 65.0 & 66.0 & 65.0 & 63.0 & 40.6 & 36.3 \\
ParaNames, six languages & \textbf{75.9} & 74.9 & 74.9 & 74.6 & 73.6 & 0.0 & 0.5 \\
GeoNames, name only & \textbf{70.1} & 68.6 & 68.7 & 69.0 & 67.6 & 0.0 & 5.5 \\
GeoNames, name with country and region & \textbf{95.1} & 92.7 & 92.3 & 94.8 & 94.8 & 32.2 & 23.4 \\
Census occupations to OCC1950 titles & 53.9 & 54.5 & \textbf{54.7} & 51.6 & 52.3 & 0.0 & 0.0 \\
Bill summaries to bill texts & \textbf{94.8} & 92.9 & 92.6 & 94.5 & \textbf{94.8} & 0.0 & 0.0 \\
Bill summaries to bill texts, 20\% unmatched & \textbf{89.9} & 88.8 & 89.5 & 87.3 & 84.1 & 0.0 & 0.0 \\
Wikipedia articles, six languages to English & \textbf{99.5} & 94.4 & 94.4 & 99.0 & \textbf{99.5} & 0.0 & 0.0 \\
Wikipedia articles, six languages to English, 20\% unmatched & 98.4 & \textbf{98.6} & \textbf{98.6} & 96.3 & 88.7 & 0.0 & 0.0 \\
\midrule
Mean & \textbf{85.2} & 83.4 & 83.6 & 83.7 & 76.0 & 31.7 & 30.6 \\
Mean loss vs.\ task-specific & --- & 1.8 & 1.6 & \textbf{1.5} & 9.3 & 2.8 & 3.9 \\
Largest loss & --- & \textbf{5.2} & \textbf{5.2} & 6.9 & 43.0 & 26.2 & 26.2 \\
\bottomrule
\end{tabular}

%% file: tables/tab_model_swap.tex
\begin{tabular}{lrrrrrrl}
\toprule
 & \multicolumn{6}{c}{Change in validation F1 (points)} & \\
\cmidrule(lr){2-7}
Configuration & Mean & NC & Names & MaDI & Occupations & Paragraphs & License added \\
\midrule
Jina v2 + BGE (baseline) & 0.0 & 0.0 & 0.0 & 0.0 & 0.0 & 0.0 & --- \\
Rank-average confidence (adopted) & +1.2 & 0.0 & +1.7 & +1.1 & +0.4 & +1.6 & --- \\
\addlinespace
Jina v2 + BGE + zerank-2 (not adopted) & +1.1 & 0.0 & +2.4 & +0.6 & +1.5 & +0.4 & Apache-2.0 \\
Jina v2 + BGE + gte-multilingual-reranker & +0.5 & 0.0 & +0.3 & +0.7 & $-$0.2 & +1.2 & Apache-2.0 \\
Jina v2 + BGE + mxbai-rerank-base-v2 & +0.2 & $-$0.2 & +0.5 & 0.0 & +0.6 & +0.2 & Apache-2.0 \\
\addlinespace
BGE + zerank-2 (no Jina v2) & +0.7 & $-$0.5 & +1.9 & +0.2 & +2.0 & 0.0 & Apache-2.0 \\
BGE + gte-multilingual-reranker (no Jina v2) & $-$0.5 & $-$0.4 & $-$1.7 & $-$1.0 & +0.9 & +0.7 & Apache-2.0 \\
BGE + mxbai-rerank-base-v2 (no Jina v2) & $-$0.9 & $-$0.8 & $-$1.7 & $-$0.9 & +0.6 & $-$0.6 & Apache-2.0 \\
\addlinespace
zerank-2 alone & +0.9 & $-$0.5 & +2.1 & +0.3 & +0.5 & --- & Apache-2.0 \\
Qwen3-Reranker-4B alone & $-$1.0 & $-$0.5 & +0.4 & $-$5.1 & +1.1 & --- & Apache-2.0 \\
jina-reranker-v3 alone & $-$1.0 & $-$0.7 & $-$3.8 & +2.5 & +2.6 & --- & CC-BY-NC-4.0 \\
KaLM-Reranker-Nano alone & $-$5.2 & $-$0.6 & $-$8.3 & $-$6.0 & $-$0.1 & --- & Apache-2.0 \\
LAMAR-600m alone & $-$0.4 & $-$0.7 & $-$0.2 & $-$1.3 & +1.1 & --- & MIT \\
GLiNER2.5-Decide alone & $-$9.4 & $-$0.5 & $-$17.0 & $-$6.9 & $-$1.7 & --- & Apache-2.0 \\
\addlinespace
Embedding: Qwen3-Embedding-4B & $-$2.5 & +0.1 & $-$5.7 & +0.4 & $-$0.6 & --- & Apache-2.0 \\
Embedding: Nemotron-3-Embed-1B & +0.6 & $-$0.1 & +1.4 & +0.3 & $-$0.4 & --- & OpenMDW-1.1 \\
\bottomrule
\end{tabular}

%% file: tables/tab_linkage_ablation.tex
\begin{tabular}{lccc}
\toprule
Configuration & NC voters & ParaNames & GeoNames \\
\midrule
Reranker ensemble alone & 93.6 & 71.8 & 79.0 \\
Dense similarity alone & 93.6 & 67.4 & 73.1 \\
Character n-gram similarity alone & 96.5 & 29.4 & 49.6 \\
Jaro-Winkler alone & 91.6 & 31.0 & 47.4 \\
\hdashline
Uniform-weight fusion & \textbf{97.8} & 73.2 & 80.8 \\
Agreement-weighted fusion & \textbf{97.8} & \textbf{73.6} & \textbf{81.2} \\
\bottomrule
\end{tabular}

%% file: tables/tab_linkage_errors.tex
\begin{tabular}{lrrrr}
\toprule
Benchmark & Queries & Not retrieved & Ranked below a wrong record & Tied \\
\midrule
NC voters, address changes & 2,000 & 3 & 42 & 1 \\
ParaNames & 18,000 & 1,417 & 3,327 & 621 \\
GeoNames & 12,000 & 626 & 1,629 & 421 \\
\bottomrule
\end{tabular}

%% file: tables/tab_calibration.tex
\begin{tabular}{lccc}
\toprule
Method & Discrimination (AUC) & ECE, raw & ECE, recalibrated \\
\midrule
\multicolumn{4}{l}{\emph{NC voters, address changes}} \\
EnsembleLink & 0.97 & --- & \textbf{0.00} \\
fastLink & 0.95 & 0.02 & \textbf{0.00} \\
Splink & \textbf{0.98} & \textbf{0.01} & 0.01 \\
fuzzylink & 0.96 & 0.07 & 0.01 \\
\multicolumn{4}{l}{\emph{ParaNames, six languages}} \\
EnsembleLink & \textbf{0.87} & --- & \textbf{0.01} \\
fuzzylink & 0.76 & 0.25 & \textbf{0.01} \\
\bottomrule
\end{tabular}

%% file: tables/tab_scaling.tex
\begin{tabular}{lrrrrrr}
\toprule
Target records & Queries & Index & Retrieve & Rerank & Total & Top-1 (\%) \\
\midrule
10,000 & 1,000 & 1.2 & 0.0 & 0.6 & 1.9 & 99.9 \\
30,000 & 1,000 & 1.3 & 0.0 & 0.6 & 2.0 & 99.7 \\
100,000 & 1,000 & 2.0 & 0.0 & 0.6 & 2.7 & 99.7 \\
300,000 & 1,000 & 3.6 & 0.1 & 0.6 & 4.3 & 99.1 \\
1,000,000 & 1,000 & 9.4 & 0.2 & 0.6 & 10.2 & 97.6 \\
100,000 & 100 & 2.0 & 0.0 & 0.1 & 2.1 & 100.0 \\
100,000 & 300 & 2.0 & 0.0 & 0.2 & 2.2 & 99.3 \\
100,000 & 3,000 & 2.0 & 0.1 & 1.8 & 3.9 & 99.3 \\
\bottomrule
\end{tabular}

%% file: tables/tab_prompt_wording.tex
\begin{tabular}{lccccc}
\toprule
Prompt & Gem. & 4o & DS4 & GLM5 & Q3.5 \\
\midrule
\multicolumn{6}{l}{\emph{Accuracy (\%)}} \\
Default & 80.7 & 74.9 & \textbf{75.4} & \textbf{75.7} & \textbf{74.9} \\
Minimal & 77.2 & 7.2 & 35.2 & 66.7 & 37.5 \\
Detailed & 79.3 & \textbf{76.5} & 75.1 & 75.0 & 74.5 \\
No expert framing & \textbf{80.8} & 69.3 & 75.1 & 73.9 & 72.3 \\
Five examples & 79.8 & 47.8 & 71.1 & 71.0 & 68.1 \\
Ten examples & 79.3 & 55.7 & 68.0 & 72.8 & 68.4 \\
\midrule
\multicolumn{6}{l}{\emph{Responses outside the five categories (of 2{,}000)}} \\
Default & 0 & 0 & 0 & 0 & 0 \\
Minimal & 115 & 1{,}842 & 1{,}083 & 262 & 1{,}108 \\
Detailed & 0 & 0 & 0 & 0 & 0 \\
No expert framing & 1 & 147 & 0 & 41 & 0 \\
Five examples & 0 & 859 & 12 & 2 & 241 \\
Ten examples & 0 & 568 & 12 & 1 & 95 \\
\bottomrule
\end{tabular}